\documentclass[10pt,twocolumn,letterpaper]{article}

\usepackage[pagenumbers]{cvpr}

\usepackage{cuted}

\usepackage{currfile}

\usepackage{caption}

\newcommand{\coderelease}{are available at \url{https://github.com/gezhixuan/zoomdecomp}}

\usepackage{multirow}
\usepackage{xspace}
\newcommand{\Iseen}{\ensuremath{I_{\mathrm{seen}}}\xspace}
\newcommand{\Isrc}{\ensuremath{I_{\mathrm{src}}}\xspace}
\definecolor{regone}{HTML}{86B6EF}\definecolor{regtwo}{HTML}{5598E7}\definecolor{regthree}{HTML}{2A78D6}\definecolor{regfour}{HTML}{184F95}
\newcommand{\regchip}[1]{\textcolor{#1}{\rule[-0.05ex]{0.8ex}{0.8ex}}\kern0.22em}
\newcommand{\rone}{\regchip{regone}R1\xspace}\newcommand{\rtwo}{\regchip{regtwo}R2\xspace}\newcommand{\rthree}{\regchip{regthree}R3\xspace}\newcommand{\rfour}{\regchip{regfour}R4\xspace}
\renewcommand{\paragraph}[1]{\vspace{.4em}\noindent\textbf{#1.}}

\makeatletter
\DeclareRobustCommand{\cond}[1]{{\sffamily\bfseries\upshape\fontsize{\dimexpr\f@size pt*9/10\relax}{\f@baselineskip}\selectfont #1}}
\makeatother

\newsavebox{\fittablebox}
\newcommand{\fittable}[1]{\sbox{\fittablebox}{#1}%
  \typeout{FITTABLE natural=\the\wd\fittablebox\space line=\the\linewidth}%
  \ifdim\wd\fittablebox>\linewidth \resizebox{\linewidth}{!}{\usebox{\fittablebox}}\else\usebox{\fittablebox}\fi}

\definecolor{cvprblue}{rgb}{0.21,0.49,0.74}
\usepackage[pagebackref,breaklinks,colorlinks,allcolors=cvprblue]{hyperref}

\def\confName{CVPR}
\def\confYear{2027}

\title{What Does a Zoom Actually Add?\\Decomposing Fine-Grained Perception Failures in Vision-Language Models}

\author{Zhixuan Ge \qquad Anqi Li \qquad Kexin Geng \qquad Wei Qiu\textsuperscript{\dag}\\
Rice University\\
{\tt\small \{zg33, al318, kg111, wq8\}@rice.edu}\\
{\small \textsuperscript{\dag}Corresponding author}}

\begin{document}
\maketitle
\begin{strip}
\centering
\includegraphics[width=\linewidth]{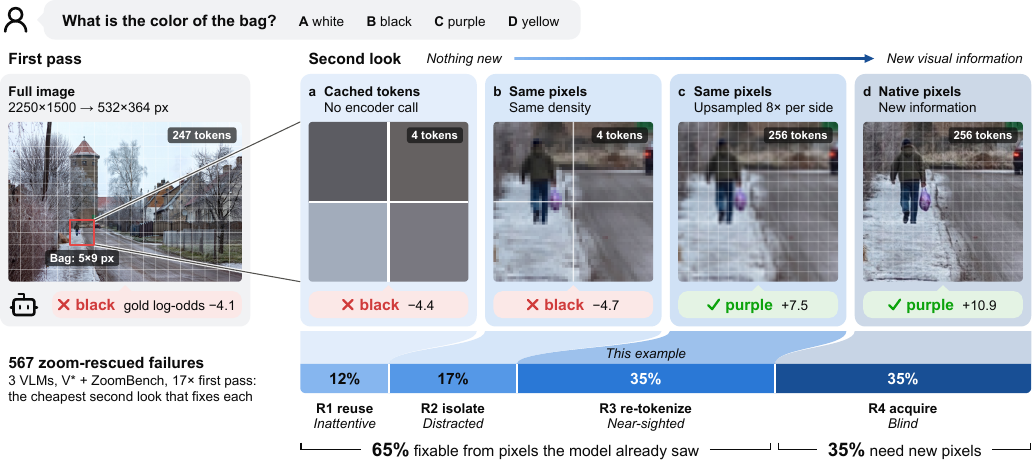}
\captionsetup{hypcap=false}
\captionof{figure}{\textbf{What does a zoom add?} \emph{Top: one zoom-rescued failure (Qwen2.5-VL-7B, V$^*$).} The first pass is fixed (left, 17$\times$ fewer pixels than the source) and only the second look varies: (a) the region's cached visual tokens (drawn as their mean colors), (b) its seen pixels re-encoded at the first-pass density, (c) the \emph{same} pixels upsampled $8\times$ per side, and (d) native-resolution pixels. White lines mark the token grid, and each answer comes with the log-odds of the correct one. The purple is already visible in (b), but the model reads it only in (c), from the same pixels on a finer grid. \emph{Bottom: all 567 zoom-rescued V$^*$ and ZoomBench failures of three VLMs at this first-pass budget.} Each failure is assigned to the cheapest second look that fixes it while its sham, the same operation on a random region of the same size, does not.}
\captionsetup{hypcap=true}
\label{fig:teaser}%
\end{strip}

\begin{abstract}
Zooming into a region often turns a wrong answer from a vision-language model (VLM) into a right one, and a growing literature takes this as evidence that the model lacked the pixels. But a zoom changes several things at once. It removes the surrounding context, re-tokenizes, at a higher density, pixels the model has already seen, and, if the first pass was downsampled, exposes detail the model never received. We separate these factors for each failure. Holding the first pass fixed, we vary only the second look and assign every failure to the cheapest one that fixes it: reusing cached tokens, isolating the region, re-tokenizing the seen pixels, or acquiring new ones. The study covers 2{,}753 zoom-rescued failures of Qwen2.5-VL, Qwen3-VL and InternVL3 on V$^*$, ZoomBench and TextVQA, at first-pass budgets from 17$\times$ fewer pixels than the source to native resolution. First, most rescued failures did not lack the pixels: upsampling the pixels the model had already seen fixes 47--73\% of them even at 17$\times$ downsampling, against 1--3\% for a random region. Second, the first-pass budget decides how often new pixels are needed: depending on the benchmark, new pixels are required for 23--50\% of the rescues at 17$\times$ and for 12--23\% at 4$\times$. The missing-pixel reading and the accounts that dispute it thus hold at different budgets. Third, the useful magnification is large and differs between objects and text: seen pixels fix more failures up to 6--12$\times$ their first-pass size per side on objects and saturate by 4$\times$ on text, and even native pixels gain from being rendered beyond native resolution. Allowing any magnification, 68--94\% of the failures at 17$\times$ need no new pixels. These models more often need a better rendering of evidence they already have than new pixels, which motivates re-tokenizing a region before acquiring it.
\end{abstract}

\section{Introduction}
\label{sec:intro}

Is a vision-language model (VLM) that misses a small detail near-sighted, distracted, inattentive, or blind? The question arises whenever a zoom helps: a VLM that answers a fine-grained question wrongly on the full image will frequently answer it correctly after a zoom, in which the relevant region is cropped out of the image and the crop is rendered with more visual tokens than the region occupied on the first pass. This observation underlies an active line of work, from training-free crop selection~\cite{vicrop2025,zoomeye2025} and learned zoom policies~\cite{deepeyes2025,minio3_2025,pixelreasoner2025,visionrl2_2026} to methods that revisit already-computed visual tokens instead of encoding anything new~\cite{sieve2026,thinkingonce2026}. They rest on different beliefs about \emph{why} the zoom helps. One account holds that the first visual pass simply did not preserve the detail~\cite{lastvisiblepixel2026,visionthink2025}. A second holds that the detail was there but too small, since enlarging the same pixels already helps~\cite{s2_2024,dragonfly2024,perceptual2024}. A third holds that zooming works mostly by removing background interference~\cite{hide2025}. A fourth holds that the evidence is already present in the tokens the model computed, so that the zoom helps only by making the model use what it has, and routing those tokens to the answer would do as well~\cite{sieve2026,thinkingonce2026,hiddenplainsight2025}. Each account is backed by gains in benchmark averages, which cannot tell for a given failure which account applies, and each implies a different system design.

We argue that the account has to be chosen per failure, not per benchmark, and that it can be. A successful zoom changes several variables at once (\cref{fig:teaser}). It removes the surrounding context and runs the vision encoder again. It re-tokenizes the target, re-encoding the same pixels at a higher density so that they span more visual tokens. Whenever the first pass was downsampled, it may also expose pixels the model never saw. These factors can be separated, and the four accounts disagree about which of them does the work. To measure what a zoom actually adds, we take each failure that an oracle zoom fixes, hold its first pass constant, and vary only what the second look adds. Each failure is assigned to the cheapest of four second looks that fixes it while a \emph{sham}, the same operation on a random region of the same size, does not. \emph{Reuse} (\rone) gives the model the region's cached visual tokens, with no new encoding: the model was inattentive. \emph{Isolation} (\rtwo) removes the background, or re-encodes the region's pixels at the density the model already saw: the model was distracted. \emph{Re-tokenization} (\rthree) upsamples the \emph{same} seen pixels, which adds token density but no information: the model was near-sighted. \emph{Acquisition} (\rfour) supplies native-resolution pixels: the model was blind. The four accounts above are these regimes in reverse order, and a native zoom cannot tell the first two apart, because it adds pixels and tokens at once. We also measure the prescription: how much the region must be magnified before the seen pixels become usable.

The decisive control is the distinction between the original image \Isrc and the exact raster \Iseen the encoder consumed on the first pass. Modern VLMs preprocess images under a pixel budget, so we sweep the first-pass budget from 17$\times$ fewer pixels than the source to native resolution, which turns ``was the information there?'' into a measured quantity rather than an assumption. We reconstruct \Iseen bit-exactly from the model's own preprocessing and build every intervention from it.

We apply this design to three open 7--8B VLMs with different tokenization policies, Qwen2.5-VL, Qwen3-VL and InternVL3, on three kinds of evidence (V$^*$ attribute and spatial questions, ZoomBench, and TextVQA text regions), using ground-truth regions as an oracle so that localization is not confounded with recognition. Three findings emerge:
\begin{itemize}
\item \textbf{Most rescued failures did not lack the pixels.} Upsampling the pixels the model had already seen fixes 47--73\% of zoom-rescued failures even at 17$\times$ downsampling, against 1--3\% for its sham, and cached tokens alone fix 4--19\%.
\item \textbf{The mix of regimes is set by the first-pass budget and the kind of evidence.} Depending on the benchmark, new pixels are needed for 23--50\% of the rescues at the 17$\times$ budget and for 12--23\% at 4$\times$. Each doubling of the downsampling multiplies the odds of needing them by 1.5--1.6 after controlling for target size. On the object benchmarks, target size predicts instead whether re-tokenization suffices.
\item \textbf{The prescription is strong.} On objects, the seen pixels become more usable up to 6--12$\times$ their first-pass size per side and then plateau, and even native pixels gain from being rendered beyond native resolution. On text the useful magnification saturates by 4$\times$, and larger renderings hurt. Allowing any magnification, 68--94\% of the rescued failures at 17$\times$ need no new pixels.
\end{itemize}
The partition also serves as a resource map for zoom policies (\cref{sec:buys}). The code and the per-sample outputs of every condition \coderelease.

\section{Related Work}
\label{sec:related}

\paragraph{Zooming as a method} Training-free cropping guided by the model's own attention~\cite{vicrop2025}, tree-structured visual search~\cite{zoomeye2025,vstar2024}, and reinforcement-learned crop tools~\cite{deepeyes2025,minio3_2025,pixelreasoner2025} all show large gains on high-resolution benchmarks~\cite{vstar2024,hrbench2025,zwz2026}. Region-level policy optimization~\cite{visionrl2_2026} finds that localization tolerates several times more token compression than recognition and re-encodes only the selected region, and VisionThink~\cite{visionthink2025} learns when to request a higher-resolution image. These methods show that a zoom helps, not what it adds. CounterCredit~\cite{countercredit2026} comes closest, auditing whether a tool call was needed and whether its returned pixels were used by comparing against random same-size patches, which we adopt as our sham design. MED~\cite{med2026} separates intrinsic learning from tool-induced effects across training checkpoints. We ask a different counterfactual, for a fixed model and a call that demonstrably helped: could the benefit have come from the already-computed representation or from already-seen pixels?

\paragraph{Reuse instead of re-encoding} SIEVE~\cite{sieve2026} re-inserts region embeddings from the original encoding and trains a policy to do so, and Thinking-Once~\cite{thinkingonce2026} routes intermediate-layer evidence without a second encoder call. Both report gains on V$^*$ and HR-Bench and conclude that evidence often survives the first pass~\cite{hiddenplainsight2025,scopd2026}. Our cached-token conditions (\cond{a1}--\cond{a3} in \cref{tab:conditions}) test the simplest form of reuse, re-inserting the region's post-projector tokens, per sample, against native zoom and against shams, and find that this reuse alone is the smallest regime.

\paragraph{Resolution versus re-tokenization} Dragonfly~\cite{dragonfly2024} compared, at training time, crops made from a 336-pixel raster with crops from the native image and found the former recovered most of the gain. S$^2$~\cite{s2_2024} showed that interpolated multi-scale inputs with no additional information still help frozen encoders. HiDe~\cite{hide2025} argues that zooming works by removing background, since masking the background at fixed resolution raises accuracy and uniform upscaling does not. Can't-Find-Waldo~\cite{waldo2026} reports that whole-image upsampling without new pixels flips many answers and attributes this to tokenization artifacts. Controlled stimuli show that target size matters~\cite{lastvisiblepixel2026}, even at fixed pixel information~\cite{perceptual2024}. These results partly conflict. Our per-sample design runs analogues of background masking, same-density re-encoding, region upsampling and native pixels on the same failures (\cond{b1}, \cond{b2}, \cond{c} and \cond{d} in \cref{tab:conditions}), and finds that the first-pass downsampling factor determines which of them is decisive.

\paragraph{Mechanistic accounts of VLM failures} Probing and patching studies trace how VLMs route visual information~\cite{pathways2026,spatialbinding2026,dualpathway2026,faithfulness2026} and find that it can be decodable yet unused~\cite{hiddenplainsight2025,vlmsblind2024,eas2026,arbitration2026,needwords2026}. Our study operates on inputs and cached tokens rather than internal activations, so every second look we test can be deployed at inference time.

\section{Decomposing a Zoom}
\label{sec:method}

\begin{table*}[t]\centering\footnotesize\setlength{\tabcolsep}{3.8pt}
\caption{\textbf{The eight second-look conditions.} crop: the pixels inside $R$. mask: the pixels outside $R$ replaced by a constant color (Appendix~\ref{sec:supp:impl}). $\downarrow$: resized to the \cond{b2} crop's size. $@K$: rendered with $K$ tokens, upsampled or downsampled as needed (the sweep of \cref{sec:magnification} replaces $K$ by $m^2 n_R$). Isolates: the one factor that differs from the bracketed condition, where \cond{a1} and \cond{b1} are compared with the first pass itself. Token counts are in units of $28^2$ pixels and approximate for Qwen3-VL (32-pixel tokens) and InternVL3 (whole tiles), see Appendix~\ref{sec:supp:impl}.}
\label{tab:conditions}
\fittable{\begin{tabular}{@{}llccllll@{}}
\toprule
 & & Visual & Encoder & Pixels & Positions & & \\
Condition & Input & tokens & call & of $R$ & of $R$ & Regime & Isolates (vs.) \\
\midrule
\cond{a1} cached re-insertion & $[\Iseen,\ \mathrm{cache}(R)]$ & $N+n_R$ & no & cache & new block & \rone & cached block added (first pass) \\
\cond{a2} region only, new positions & $[\mathrm{cache}(R)]$ & $n_R$ & no & cache & new block & \rone & full image dropped (\cond{a1}) \\
\cond{a3} region only, original positions & $[\Iseen]$ minus background tokens & $n_R$ & no & cache & original & \rone & positions (\cond{a2}) \\
\cond{b1} background masked & $[\mathrm{mask}(\Iseen, R)]$ & $N$ & yes & seen & original & \rtwo & background masked (first pass) \\
\cond{b2} seen crop, same density & $[\Iseen,\ \mathrm{crop}(\Iseen, R)]$ & $N+n_R$ & yes & seen & new block & \rtwo & cached vs.\ re-encoded (\cond{a1}) \\
\cond{b3} native crop, same density & $[\Iseen,\ \mathrm{crop}(\Isrc, R)\!\downarrow]$ & $N+n_R$ & yes & native & new block & control & pixel source (\cond{b2}) \\
\cond{c} seen crop, upsampled & $[\Iseen,\ \mathrm{crop}(\Iseen, R)\,@K]$ & $N+K$ & yes & seen & new block & \rthree & token density (\cond{b2}) \\
\cond{d} native crop, oracle & $[\Iseen,\ \mathrm{crop}(\Isrc, R)\,@K]$ & $N+K$ & yes & native & new block & oracle & pixel source (\cond{c}) \\
\bottomrule
\end{tabular}}
\end{table*}

\paragraph{Setting} A VLM turns an image into a grid of \emph{visual tokens} through a vision encoder and a projector, and the language model receives one token block per image followed by the question. Every condition below is a new prompt, answered from scratch. Before encoding, every VLM resizes the image under a pixel budget $B$, and \Iseen\ is the exact raster the encoder consumed, reconstructed from the source image \Isrc\ with the model's own processor (Appendix~\ref{sec:supp:impl}). The first-pass \emph{downsampling factor} $|\Isrc|/|\Iseen|$, our main experimental variable, is a ratio of pixel counts: a $17\times$ reduction in pixel count corresponds to roughly a $4\times$ downscaling per side. The first pass yields $N$ visual tokens, and the target region $R$ (the ground-truth box, padded and snapped to the token grid) spans a $g_w\times g_h$ block of $n_R=g_w g_h$ of them. For a fixed region, $n_R$ measures the \emph{token density} at which it is encoded. The oracle zoom (condition \cond{d} in \cref{tab:conditions}) appends a crop of \Isrc\ at $R$ to the first-pass image. A \emph{zoom-rescued failure} is a sample that the first pass answers wrongly and the oracle zoom answers correctly, made precise under \emph{Scoring} below. We take the region as given, since localization has its own failure mechanisms~\cite{visionrl2_2026}.

\paragraph{Second looks} On every zoom-rescued failure we run the eight conditions of \cref{tab:conditions}, each a list of token blocks for the language model (\cref{fig:teaser}, top, with further examples in \cref{fig:examples}). They follow the order of \cref{sec:intro}, from reuse to new pixels, and each differs from a cheaper one in a single factor (last column of \cref{tab:conditions}). $\mathrm{cache}(R)$ is the language model's input for the $n_R$ tokens of $R$: the projector output saved from the first pass, so the encoder is skipped while the language model runs anew. In \cond{a1} and \cond{a2} these vectors receive the positional encodings of a new image block, and \cond{a3} keeps the ones they had in the first pass (image row and column for the Qwen models, sequence index for InternVL3) by deleting the background tokens from the first-pass sequence. The cached vectors were computed with the whole image in view, so \cond{a2} and \cond{a3} remove the background only inside the language model, whereas \cond{b1} and \cond{b2} remove it at the encoder, by masking it (\cond{b1}) or cropping it away (\cond{b2}).

\paragraph{Crop budget and magnification} Conditions \cond{c} and \cond{d} render the crop with a budget of $K=\min(B,\max(256, |R|))$ tokens, where $B$, $K$ and the region's size in the source image $|R|$ are all counted in units of $28^2$ pixels. The budget is the region's native size, raised to at least 256 tokens so that small regions remain legible and capped at the first-pass budget, and the processor rounds the crop to its token grid (Appendix~\ref{sec:supp:impl}). Condition \cond{d} therefore shows source pixels at about $K$ tokens, which is above native resolution for a small region and below it for a region larger than the budget. Both conditions magnify the region by $m=\sqrt{K/n_R}$ per side, which varies with the region's size. For example, a $448\times448$ \Iseen\ has $N=256$ tokens, a $3\times3$-token region has $n_R=9$, and at $K=256$ \cond{c} renders it at $m\approx5.3$ times its first-pass side. The \emph{magnification sweep} (\cref{sec:magnification}) instead renders every region at a fixed $m$ times its first-pass side, i.e.\ with $m^2 n_R$ tokens, so $m=1$ reproduces \cond{b2} and its native-pixel control \cond{b3}, and native resolution corresponds to $m\approx4$ at $17\times$.

\paragraph{Shams} Each condition has a \emph{sham}: the same operation on a random region $R'$ of the same token-grid size, placed to have zero overlap with $R$ where possible. A sham captures everything the condition changes apart from the target: sequence length, a second block, re-encoding, and chance flips.

\paragraph{Scoring} For multiple-choice questions we read the log-probabilities of the option letters at the first answer position and report the \emph{gold log-odds} $\log p(\text{gold}) - \log\sum_{o\neq\text{gold}}p(o)$. For counting questions we use the first digit and for free-text answers the first token of the gold string (Appendix~\ref{sec:supp:impl}). Greedy correctness is exact match for letters and digits and VQA soft accuracy $\geq0.5$ for text. A condition \emph{fixes} a sample only if its greedy answer is correct \emph{and} its gold log-odds is positive. The second clause removes chance flips. A \emph{failure} is a sample whose first-pass greedy answer is wrong, whatever its log-odds, and a zoom-rescued failure is a failure that the oracle zoom \cond{d} fixes while its sham does not, so that the rescue is attributable to the target.

\paragraph{Regimes} The four regimes are ordered by cost: no encoder call (\rone\ reuse, \cond{a1}--\cond{a3}), re-encoding at first-pass density (\rtwo\ isolation, \cond{b1} or \cond{b2}), re-tokenization of seen pixels (\rthree, \cond{c}) and new pixels (\rfour\ acquisition). Each rescued failure goes to the cheapest regime with a condition that fixes it while its sham does not. This is the \emph{strict} partition, and the \emph{lenient} one drops the sham requirement. The split is by cost, not by mechanism: a fix that drops the background inside the language model (\cond{a2}, \cond{a3}) needs no encoder call and counts as \rone. Condition \cond{b3} is a control outside the partition (\cref{sec:exp}), \cond{d} fixes every sample by definition of the cohort, and \rfour\ is the residual. The composition of \rfour, the lenient partition and violations of the hierarchy are in Appendices~\ref{sec:supp:cohort}, \ref{sec:supp:lenient} and~\ref{sec:supp:robust}.

\begin{figure*}[t]\centering
\includegraphics[width=\linewidth]{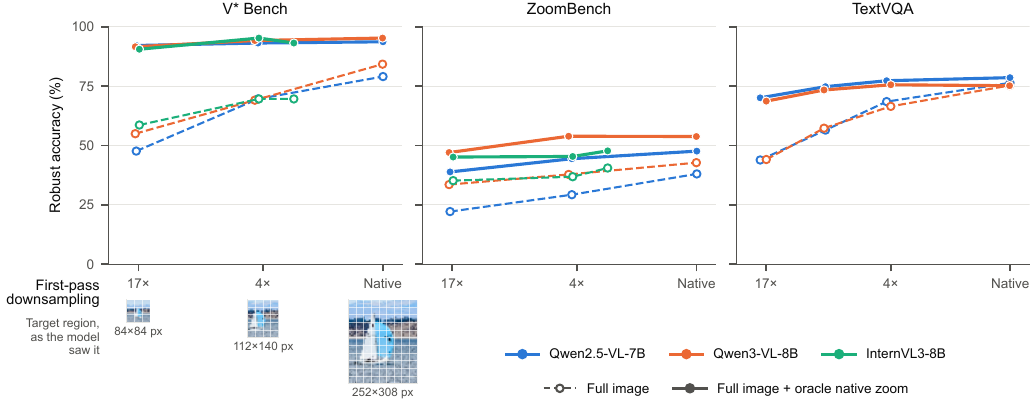}
\caption{\textbf{First-pass budget and oracle zoom.} Robust accuracy (greedy answer correct and gold log-odds positive) of the full image at a given first-pass downsampling (dashed, hollow markers) and with the oracle native zoom appended (solid), for three models and three benchmarks (native: the finest budget). Under the V$^*$ axis: the target region of one V$^*$ image as the model saw it at each budget, at true relative size with its token grid. Zoom rescues a large, budget-dependent slice, and on V$^*$ it is near ceiling everywhere.}
\label{fig:budget}
\end{figure*}

\begin{figure*}[t]\centering
\includegraphics[width=\linewidth]{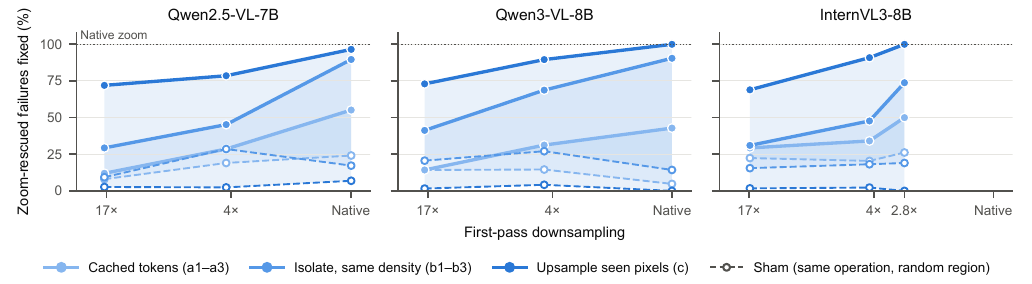}
\caption{\textbf{What a cheaper intervention recovers.} For each model, the share of zoom-rescued V$^*$ failures also fixed by at least one condition of each family as the first pass gets finer. Dashed hollow markers are the shams, shaded areas the effect beyond the sham. Upsampling already-seen pixels fixes most rescues even at 17$\times$, and reuse and isolation become competitive only near native resolution, where the seen and native crops coincide.}
\label{fig:sweep}
\end{figure*}

\paragraph{Models, benchmarks and budgets} We study three open VLMs that differ in how pixels become tokens: Qwen2.5-VL-7B-Instruct~\cite{qwen25vl2025} (dynamic resolution, 28 pixels per token), Qwen3-VL-8B-Instruct~\cite{qwen3vl2025} (32 pixels per token and multi-level ``DeepStack'' features, which we cache and inject alongside the main tokens) and InternVL3-8B~\cite{internvl3_2025} (448-pixel tiles of 256 tokens plus a thumbnail). The benchmarks are V$^*$~\cite{vstar2024} (191 attribute and spatial questions with target boxes), ZoomBench~\cite{zwz2026} (844 of its 845 multiple-choice and counting questions with key regions) and 1{,}000 random questions of the TextVQA~\cite{textvqa2019} subset with text regions released with ViCrop~\cite{vicrop2025}, scored by VQA soft accuracy. Budgets are set in pixels, so that the first-pass downsampling factor is matched across models. On V$^*$ and ZoomBench, $B\in\{0.2, 0.8, 3.2, 12.8\}$ megapixels (MP) gives median downsampling factors of 17$\times$, 4$\times$, 1.1$\times$ and 1.0$\times$. InternVL3's tile policy maps these budgets to 1, 4 and at most 12 tiles, so its finest first pass is 2.8$\times$. TextVQA images have about 1\,MP, and $B\in\{0.05, 0.1, 0.2, 0.8\}$\,MP gives 17--18$\times$, 9$\times$, 4$\times$ and native resolution. Factors differ slightly between models, and figures and tables label each cell with its own median. InternVL3 is not run on TextVQA, because one 448-pixel tile, its smallest input, is already the 4$\times$ budget there. We call one model on one benchmark at one budget a \emph{cell}. Over all cells the oracle zoom rescues 2{,}753 failures (\cref{tab:stage1full}), and the regime partition uses the 2{,}202 from cells downsampled at least $2\times$, where \cond{c} and \cond{d} differ.

\section{Is the Information Missing?}
\label{sec:exp}

\paragraph{Predictions} If zoom works because the first pass lost the pixels, then upsampling the pixels the model already received (\cond{c}) should fix no more failures than its sham, and only native pixels (\cond{d}) should help. If zoom works by re-tokenizing what the model already saw, \cond{c} should approach \cond{d}, and the gap between them should be the only part of the benefit that depends on new information.

\paragraph{What there is to explain} \Cref{fig:budget} shows the size of the effect. On V$^*$ the oracle zoom is near ceiling (91--95\%) at every budget, while the first pass climbs from 48--59\% at 17$\times$ to 70\% for InternVL3 at its finest budget and to 79--84\% for the two Qwen models at native resolution, where a zoom adds no pixels and still gains 11--15 points. ZoomBench and TextVQA have lower ceilings. Per cell, 21--75 (V$^*$), 81--151 (ZoomBench) and 19--239 (TextVQA) questions are zoom-rescued (\cref{tab:stage1full}). The zoom also hurts 3--53 questions per cell, which we set aside (Appendix~\ref{sec:supp:hurt}).

\paragraph{Result: most rescues need no new pixels} \Cref{fig:sweep} shows, per model, the share of zoom-rescued V$^*$ failures that at least one condition of each cheaper family also fixes, as the first pass gets finer, together with the shams. Upsampling the pixels the model had already seen (\cond{c}) fixes 69--73\% of them at 17$\times$, while its sham stays at or below 7\% at every budget. Isolation and same-density re-encoding (any of \cond{b1}--\cond{b3}) fix 29--41\% at 17$\times$ and most near native resolution. Cached tokens (any of \cond{a1}--\cond{a3}) stay within 7 points of their shams at 17$\times$ and help only near native resolution. Pooled over the rescued V$^*$ and ZoomBench failures (\cref{tab:arms}, intervals in \cref{tab:paired}), \cond{c} is 63--72 points above its sham and \cond{a1} 3--7 points. The same ordering holds on ZoomBench and TextVQA (\cref{fig:sweep_zb}). Two controls locate the effect. Condition \cond{b3}, the native crop resized to the \cond{b2} size, is the same image up to rounding (Appendix~\ref{sec:supp:impl}) and fixes nearly the same samples as \cond{b2}, with paired differences of at most 1.2 points and 2--4\% of discordant samples: the step from \cond{b2} to \cond{c} is a change of density, not of pixel source. Condition \cond{a3} keeps the cached region tokens at their original positions while \cond{a2} assigns them new ones, and the two agree within 2 points, so positions do not matter for reuse. What a zoom adds, beyond the share that only native pixels fix, is a better tokenization of pixels the model already had: more tokens on the target and its isolation from the background.

\begin{table}[t]\centering\footnotesize\setlength{\tabcolsep}{2.5pt}
\caption{\textbf{Target region versus random region}, pooled over zoom-rescued V$^*$ and ZoomBench samples with a first pass downsampled $\geq2\times$ ($n=372$, 397 and 424 from top to bottom): percentage fixed by each condition on the target region (by its sham in parentheses). The rates of different conditions overlap.}
\label{tab:arms}
\fittable{\begin{tabular}{@{}lccccccc@{}}
\toprule
Model & \cond{a1} & \cond{a2} & \cond{a3} & \cond{b1} & \cond{b2} & \cond{b3} & \cond{c} \\
\midrule
Qwen2.5-VL-7B & 4 (1) & 13 (7) & 14 (6) & 23 (9) & 20 (2) & 20 (1) & 67 (1) \\
Qwen3-VL-8B & 9 (4) & 21 (14) & 21 (15) & 38 (15) & 27 (4) & 27 (4) & 65 (2) \\
InternVL3-8B & 8 (1) & 25 (14) & 27 (16) & 32 (15) & 21 (2) & 22 (2) & 74 (2) \\
\bottomrule
\end{tabular}
}
\end{table}

\section{Near-Sighted, Distracted, Inattentive, or Blind?}
\label{sec:regimes}
The previous section reported how often each condition fixes a failure, and those rates overlap. Here each failure is assigned to exactly one regime, so the shares sum to 100\%. If the four regimes are distinct rather than one effect measured at different strengths, three things should hold. Every regime should be populated for every model and benchmark. The mix should shift toward the cheaper regimes as the first pass gets finer, because a finer first pass leaves fewer pixels unseen. And whether a failure needs new pixels should depend on the first-pass budget rather than on how large the target is.

\begin{figure*}[t]\centering
\includegraphics[width=\linewidth]{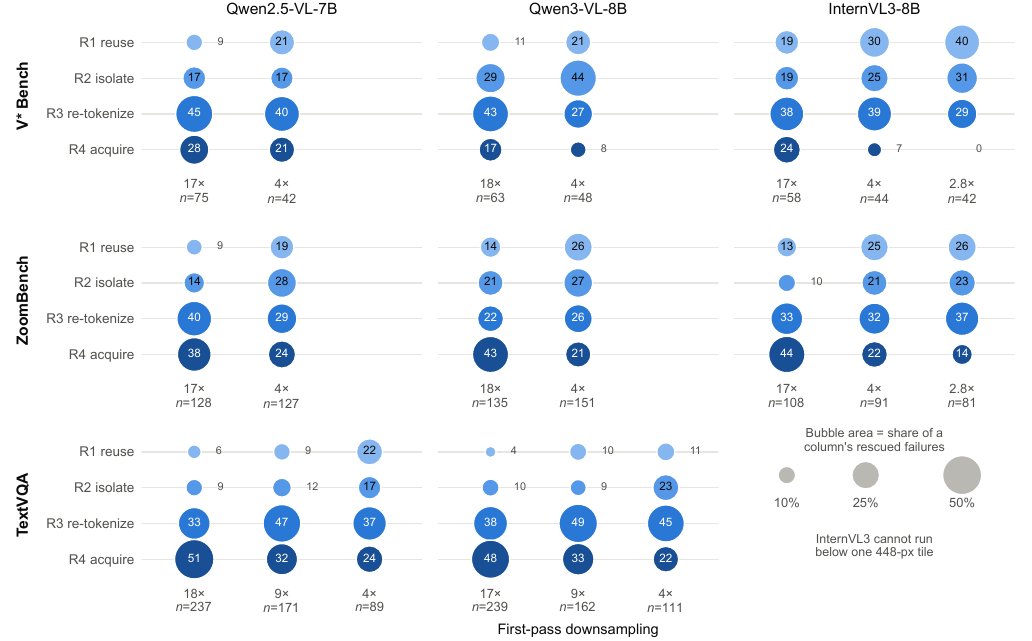}
\caption{\textbf{Strict regime partition of zoom-rescued failures.} Panel rows are benchmarks and panel columns are models. Within a panel, rows are regimes and columns first-pass budgets (coarse to fine), and the area of each dot is the share of that budget's rescued failures assigned to the regime (number in \%, $n$ under each column). This is the strict partition: a condition counts only if it fixes the sample while its sham does not. Reading down a column gives the partition at one budget, and reading along a row shows how a regime grows or shrinks as the first pass gets finer: \rfour\ empties, down to zero for InternVL3 on V$^*$ at 2.8$\times$, and \rone/\rtwo\ fill. Text needs new pixels most often.}
\label{fig:regimes}
\end{figure*}

\paragraph{Result: four populated regimes whose mix follows the budget} Assigning each failure to the cheapest condition that beats its sham gives the partition of \cref{fig:regimes}. All four regimes are populated for every model and benchmark. Averaged over models, \rfour\ holds 23\% of the rescues on V$^*$, 41\% on ZoomBench and 50\% on TextVQA at the 17$\times$ budget. At 4$\times$ it holds 12\%, 23\% and 23\%, and the reuse and isolation regimes have grown correspondingly (\cref{fig:nativecurve}). The ordering of regimes with budget is the same for all three models.

\paragraph{Benchmarks differ in crop size} ZoomBench regions are less magnified than V$^*$ regions at the same crop budget (a median $m$ of 4.0--4.6 at 17$\times$, against 7--8), which the sweep of \cref{sec:magnification} corrects. Text needs new pixels most often. At the 18$\times$ budget, upsampling fixes under half of the text rescues (47--50\%), against seven in ten on V$^*$, and the cached-token conditions stay within 3 points of their shams. Here the crop budget is capped at the first-pass budget of 64 tokens: the median text region is magnified 3.0--3.4$\times$, against 7--8$\times$ on V$^*$, and for most text rescues even the native crop is rendered about 1.3$\times$ below native resolution. The sweep lifts the cap and still leaves 27--35\% of text rescues out of reach of the seen pixels, against 6--17\% on V$^*$.

\paragraph{One axis for prior results} The budget places the competing accounts on one axis. Near native resolution, isolation and reuse dominate, as in HiDe's masking result~\cite{hide2025}. At moderate downsampling, re-tokenization dominates, as in Dragonfly's upsampling result~\cite{dragonfly2024}. At strong downsampling and for text, \rfour\ holds up to 50\% of the rescues, as the missing-pixel account~\cite{lastvisiblepixel2026} expects.

\paragraph{Result: budget decides blindness} \Cref{fig:phase} plots each regime's share against the number of first-pass tokens the target occupied. \rthree\ holds 42\% of the failures with targets below half a token and 27\% of those above eight tokens, reuse rises with target size, and the \rfour\ share shows no trend. A logistic regression with benchmark and model fixed effects (\cref{tab:logit}) separates the two variables. Each doubling of the first-pass downsampling multiplies the odds of needing new pixels by 1.5--1.6, an effect that is significant in the pooled fits and in each benchmark fitted alone, while target size has no effect on V$^*$+ZoomBench and a small negative one once TextVQA is included. Conversely, on the object benchmarks \rthree\ is predicted by small targets and not by downsampling. On objects, blindness is a property of the budget, and near-sightedness of the target's size. Among the 304 questions that the same model rescues at both the coarsest and the 4$\times$ first pass, the \rfour\ share falls from 45\% to 16\%. Of these questions, 93 move to a cheaper regime and 5 the other way (\cref{tab:budgetcommon}, per-question trajectories in \cref{fig:flow,fig:trajectory}), and the shift is not produced by the changing crop budget either (Appendix~\ref{sec:supp:budget}). The underlying failures persist under a paraphrased prompt, most rescues survive a wider crop, and the partition keeps its ordering under the lenient rule and under other cohort definitions (Appendices~\ref{sec:supp:cohort} and~\ref{sec:supp:robust}).

\section{How Strong a Prescription?}
\label{sec:magnification}
\begin{figure*}[t]\centering
\includegraphics[width=\linewidth]{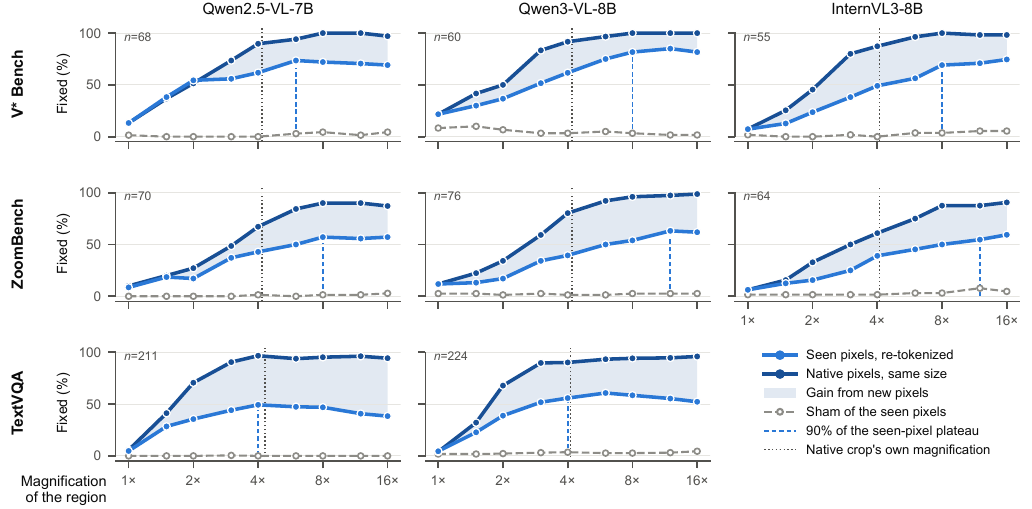}
\caption{\textbf{How much must the region be magnified?} Share of the zoom-rescued failures at the coarsest first pass that are fixed when the target region is rendered at $m$ times its first-pass size per side: from the pixels the model had already seen, from native pixels at the same output size, and from the sham of the seen pixels, on the failures whose region fits $16\times$ ($n$ per panel). The dotted line marks the native crop's own magnification ($\approx4\times$ at a 17$\times$ first pass), and the dashed marker the magnification at which the seen-pixel curve reaches 90\% of its plateau. Objects gain up to about $8\times$, native pixels included, and then plateau. Seen-pixel text saturates by $4\times$ and then declines.}
\label{fig:prescription}
\end{figure*}

If the near-sighted regime (\rthree) is real, there should be a dose-response. The seen pixels should become usable once the region is rendered large enough, the required magnification should be measurable, and native pixels should stop adding anything once their own detail is exhausted. \Cref{fig:prescription} sweeps the magnification $m$ of the region per side, from $1\times$ (\cond{b2}/\cond{b3}) to $16\times$ its first-pass size, for seen pixels, native pixels and the sham of the seen pixels. We use the zoom-rescued failures at the coarsest budget whose region can be rendered at every magnification up to $16\times$, and Appendix~\ref{sec:supp:mag} extends the sweep to $32\times$.

\paragraph{Result: objects need 6--12$\times$ magnification, text 4$\times$} On V$^*$ and ZoomBench the seen-pixel curve rises until about $8\times$ and then flattens. At $8\times$ it fixes 50--82\% of the failures, against at most 4\% for its sham. The prescription is strong: these models need the target rendered at 6--12$\times$ its first-pass size per side, i.e.\ 36--144 times the tokens, before the seen pixels reach 90\% of what they can fix (dashed markers). The third expectation fails in an instructive way: native pixels keep helping past native resolution. The native crop's own detail corresponds to $m\approx4$ (dotted line), yet rendering it at $8\times$, which adds no pixels, fixes 8--27 points more. Magnification is a resource in its own right, over and above information. Text behaves differently. Its seen-pixel curve saturates by $4\times$ (49--56\%) and declines at larger sizes, losing 16--18 points from $8\times$ to $32\times$ (Appendix~\ref{sec:supp:mag}). Native pixels reach about 90\% by $3\times$ and stay there. The gap between the two curves, the value of information, does not close with magnification. Counting a failure as fixable without new pixels if a cheaper condition or some magnification of its seen pixels beats its sham, 68--94\% need no new pixels (Appendix~\ref{sec:supp:mag}). The \rfour\ share of \cref{fig:regimes} is an upper bound, which the sweep lowers by 9--23 points.

\section{What the Partition Buys}
\label{sec:buys}
Read as a resource map, the partition says that at 17$\times$ no new pixels are needed for 49--83\% of the zoom-rescued failures per model and benchmark: cached tokens alone fix 4--19\% (richer routing methods~\cite{sieve2026,thinkingonce2026} may extract more), and one more encoder call over pixels the model already has fixes the rest. Only the remainder justifies fetching new pixels. That remainder is largest for text and at low budgets.

\paragraph{A re-tokenize-first policy} \Cref{tab:policy} evaluates the simplest consequence on the existing outputs: re-tokenize the region first, and acquire native pixels only if the model's probability for its own answer after re-tokenization, $p_{\max}$, stays below a threshold. On the zoom-rescued failures at 17$\times$, stopping at $p_{\max}\geq0.9$ acquires native pixels for 36--68\% of the failures and keeps 81--98\% of the rescues. Trying cached tokens first is worse at every threshold, because their answers are more often confidently wrong (\cref{tab:policyfull}). The price is one extra encoder call for every failure whose native pixels are fetched anyway.

\paragraph{Limitations} Regions are oracle: the partition says what recognition needs once the model knows where to look, and on a 17$\times$ first pass the models' own predicted boxes cover the target for only 19--62\% of questions. With these boxes, 33--78\% of the rescued failures are fixed by no condition (Appendix~\ref{sec:supp:ground}). Regimes are defined by the eight conditions tested, and the cohort definition is one of several defensible choices (Appendix~\ref{sec:supp:cohort}). Each sample carries one label from one crop and prompt. Under a second, independent placement of every sham, 95\% of the object failures and 99\% of the text failures keep their label, and each model's regime shares pooled over budgets move by less than 2 points (Appendix~\ref{sec:supp:robust}). The benchmarks favor small, nameable targets, and all three models are 7--8B.

\begin{table}[t]\centering\footnotesize
\caption{\textbf{Rescues kept and native pixels acquired by a re-tokenize-first policy} on the zoom-rescued V$^*$+ZoomBench failures at 17$\times$, in percent of these failures, when the policy stops at $p_{\max}\geq0.9$. Always acquiring keeps every rescue, and each acquisition costs one more encoder call (\cref{tab:policyfull}).}
\label{tab:policy}
\fittable{\begin{tabular}{@{}lccc@{}}
\toprule
 & Re-tokenize only & \multicolumn{2}{c}{Re-tokenize, then acquire} \\
\cmidrule(lr){2-2}\cmidrule(l){3-4}
Model & rescues kept & acquisitions & rescues kept \\
\midrule
Qwen2.5-VL-7B & 63 & 68 & 98 \\
Qwen3-VL-8B & 55 & 36 & 81 \\
InternVL3-8B & 58 & 61 & 93 \\
\bottomrule
\end{tabular}
}
\end{table}

\section{Conclusion}
\label{sec:conclusion}
We asked what a zoom contributes to the failures it fixes and answered it per failure, with eight conditions and their shams on 2{,}753 zoom-rescued failures of three VLMs across a sweep of first-pass budgets. First, most rescued failures did not lack the pixels: even at 17$\times$ downsampling, upsampling the pixels the model had already seen fixes 47--73\% of them. Second, the mix of regimes is set by the first-pass budget and the kind of evidence, so the accounts of zooming describe regions of one axis rather than rivals. Third, the useful magnification is large, 6--12$\times$ per side on objects and 4$\times$ on text, and at the right magnification most rescued failures need no new pixels. Once the region is known, a zoom policy should therefore re-tokenize before it acquires.

\section*{LLM Usage Statement}
Claude (Anthropic) wrote parts of the code under the authors'
direction, and the authors reviewed and tested it. Claude also
assisted with the analysis of results, figure scripts, literature
search, discussion of ideas and refinement of text written by the
authors. Claude and ChatGPT (OpenAI) gave reviewer-style feedback on
drafts. The vision-language models evaluated in our experiments are
objects of study and are described in the main text. The authors
verified all results and take full responsibility for the content.

{
    \small
    \bibliographystyle{ieeenat_fullname}
    \bibliography{main}
}

\clearpage
\appendix
\setcounter{figure}{0}\renewcommand{\thefigure}{S\arabic{figure}}
\setcounter{table}{0}\renewcommand{\thetable}{S\arabic{table}}
\maketitlesupplementary
\section{Implementation details}
\label{sec:supp:impl}
\paragraph{Seen raster} For Qwen2.5-VL (28-pixel tokens: 14-pixel patches merged $2\times2$) and Qwen3-VL (32-pixel tokens: 16-pixel patches merged $2\times2$) the processor resizes an image to a multiple of the token size under $(\min,\max)$ pixel bounds. We call the processor's own resize on the uint8 image, which reproduces its pixel values exactly (a direct bicubic call differs by up to three gray levels). For InternVL3 the first pass is the stock dynamic tiling of the raw image with at most $n=\mathrm{round}(B/448^2)$ tiles ($1\le n\le12$): the processor picks the tile grid $\text{cols}\times\text{rows}$ with $\text{cols}\cdot\text{rows}\le n$ closest to the image aspect ratio, resizes the image to $448\,\text{cols}\times448\,\text{rows}$ pixels (aspect-distorting), splits it into tiles, and appends a $448^2$ thumbnail of the whole image whenever more than one tile is used. The seen canvas is this $448\,\text{cols}\times448\,\text{rows}$ raster, reconstructed by the same resize. It is the pixel source for crops and masks, and re-tiling it reproduces the first-pass tiles. The thumbnail is produced by the processor from the raw image. No condition crops from it, its tokens enter the cached-token conditions only as part of the first-pass context, and in \cond{b1} the processor builds it from the masked canvas, so it is masked too.

\paragraph{Visual tokens and indexing} We cache the post-projector visual tokens returned by the model's image-feature function, i.e.\ the tokens the language model consumes. For Qwen2.5-VL and Qwen3-VL they are in row-major order over the merged grid, so the token of grid cell $(g_x,g_y)$ has index $g_y\,G_W+g_x$. Qwen3-VL additionally returns its three DeepStack feature maps, which are aligned one-to-one with the main tokens and which we cache and inject for exactly the same token subset. For InternVL3 the tokens are tile-major (256 per tile after pixel shuffle, 16 per side, i.e.\ 28 pixels per token): cell $(g_x,g_y)$ of the canvas maps to tile $\lfloor g_y/16\rfloor\,\text{cols}+\lfloor g_x/16\rfloor$ and offset $(g_y\bmod16)\,16+(g_x\bmod16)$. The thumbnail's tokens come last and are never selected as region tokens. Cached-token conditions replace the image-feature call by the cached tensors, so no encoder runs.

\paragraph{Second looks} \cond{a1} appends a second image block whose placeholders are filled with the cached region tokens, and the language model assigns it fresh positions (multimodal RoPE for Qwen, 1-D positions for InternVL3). \cond{a2} keeps only the region tokens, as the sole image. \cond{a3} decodes greedily, like every other condition, from the original first-pass sequence with every non-region image token removed: for Qwen the 3-D position ids are computed by the model's own rope-index routine on the full sequence and then sliced to the kept positions, so each kept token retains its original $(t,h,w)$ position, and the DeepStack embeddings are sliced identically. For InternVL3 the 1-D positions are sliced the same way. \cond{b1} replaces the first image by the seen raster with everything outside the region set to the mean color (the processor's normalization mean, the color that normalizes to zero), processed exactly as the first pass (same dimensions, same tiling, same token count). \cond{b2} encodes the crop of the seen raster at the region as a second image: for Qwen its dimensions are already multiples of the token size, so the region receives exactly as many tokens as on the first pass. For InternVL3 the crop is pasted at the top-left of a mean-color canvas of $\lceil w/448\rceil\times\lceil h/448\rceil$ tiles, which the processor tiles without resizing (a thumbnail is added only when that canvas has more than one tile). \cond{b3} is the native crop resized to the \cond{b2} crop's pixel dimensions and processed identically. Since the processor's resize is applied either way, the two inputs differ only through the crop boundary and the scale rounding (on 40 V$^*$ samples at 17$\times$: mean absolute difference 1.2 gray levels, 95th percentile 4, pixel correlation 0.999), which is why \cond{b2} and \cond{b3} fix nearly the same samples. \cond{c} and \cond{d} are the seen crop and the native crop resized to the same dimensions, chosen by the processor's resize rule for the crop budget $K$ (Qwen) or to $\mathrm{round}(K/448^2)$ tiles (InternVL3). \cond{d} is therefore downsampled from \Isrc\ whenever the native region exceeds $B$, which at the coarsest budget happens for about half of the ZoomBench rescues (45--53\%), for nearly all TextVQA rescues (98--100\%) and for 6--11\% on V$^*$. \Cref{tab:dims} lists the resulting second-look sizes and token counts at the 17$\times$ budget. Across all 8{,}587 samples that received a second look, \cond{c} and \cond{d}, and likewise \cond{b2} and \cond{b3}, have identical pixel dimensions and token counts, and \cond{a2} and \cond{a3} identical token counts. For InternVL3 a sub-tile \cond{b2} crop occupies one tile, so \cond{b2} and \cond{c} have the same token count and differ only in pixel density. The floor of 256 tokens in $K$ is one 448-pixel tile and equals the coarsest first-pass budget. Token counts in \cref{tab:conditions} are in units of $28^2$ pixels, the token of Qwen2.5-VL, and hold up to the rounding of a crop to the token grid: Qwen3-VL has 32-pixel tokens and therefore $(28/32)^2$ as many, and InternVL3 works in whole tiles.

\paragraph{Regions} The union of ground-truth boxes is padded by $0.5\times$ its larger side, mapped to the first-pass token grid and snapped outward to whole tokens (at least $2\times2$). The random region of a sham has the same token-grid size and zero overlap with the target where possible.

\paragraph{Prompts and software} A prompt is the benchmark question followed by one fixed instruction line: ``Answer with the option's letter from the given choices directly.'' for multiple choice (options listed as ``A.\ \dots''), ``Answer with a single number.'' for counting and ``Answer the question using a single word or phrase.'' for TextVQA. The paraphrase check replaces this line. Each model's chat template is used unchanged, with the image(s) before the text. Software: transformers 5.12.1, torch 2.6 with CUDA 12.4, bfloat16, PyTorch SDPA attention, greedy decoding, inference only on a single H200. The random region of a sham is placed by a generator seeded with the CRC32 of ``sample id$|$budget'', so all shams of a sample share one random region per budget. The second placement of Appendix~\ref{sec:supp:robust} seeds the generator with the CRC32 of ``sample id$|$budget$|$sham2'' and reruns only the eight shams, because the target conditions do not depend on the placement. The TextVQA subsample is drawn with seed 0. One of the 845 ZoomBench items is excluded because its answer key (D) is not among its two options.

\paragraph{Scoring} Decoding is greedy with up to 8 (12 for text) new tokens, and the first-step logits give the letter, digit or first-token log-probabilities. For counting questions greedy correctness is exact match of the full generated integer, and the log-odds is that of the gold answer's first digit among the digit tokens. Text answers use VQA soft accuracy over the ten human answers: greedy correctness requires soft accuracy $\geq0.5$ (at least two annotators gave the normalized answer), and the gold log-odds is the log-odds of the first token of the majority answer, summed over its casing variants.

\begin{table*}[h]\centering\footnotesize\setlength{\tabcolsep}{3pt}
\caption{\textbf{What each condition feeds the model} at the 17$\times$ first-pass budget on V$^*$ (medians over samples): pixel size of the second look, number of image tokens in the context and total input tokens. The first pass alone has 247 / 187 / 256 image tokens for the three models. Cached-token conditions run no encoder, \cond{a2} and \cond{a3} drop the first-pass tokens, and \cond{b1} replaces the first image.}
\label{tab:dims}
\fittable{\begin{tabular}{lrrrrrrrrr}
\toprule
 & \multicolumn{3}{c}{Qwen2.5-VL-7B} & \multicolumn{3}{c}{Qwen3-VL-8B} & \multicolumn{3}{c}{InternVL3-8B} \\
\cmidrule(lr){2-4} \cmidrule(lr){5-7} \cmidrule(lr){8-10}
Condition & 2nd look (px) & image tok. & input tok. & 2nd look (px) & image tok. & input tok. & 2nd look (px) & image tok. & input tok. \\
\midrule
first pass only & -- & 247 & 303 & -- & 187 & 233 & -- & 256 & 303 \\
\cond{a1} cached re-insertion & (cached) & 251 & 311 & (cached) & 191 & 240 & (cached) & 260 & 311 \\
\cond{a2} region only, new positions & (cached) & 4 & 62 & (cached) & 4 & 51 & (cached) & 4 & 52 \\
\cond{a3} region only, original positions & (cached) & 4 & 62 & (cached) & 4 & 51 & (cached) & 4 & 52 \\
\cond{b1} background masked & full raster & 247 & 304 & full raster & 187 & 233 & full raster & 256 & 303 \\
\cond{b2} seen crop, same density & 56$\times$56 & 251 & 311 & 64$\times$64 & 191 & 240 & 56$\times$56 & 512 & 562 \\
\cond{b3} native crop, same density & 56$\times$56 & 251 & 311 & 64$\times$64 & 191 & 240 & 56$\times$56 & 512 & 562 \\
\cond{c} seen crop, upsampled & 448$\times$448 & 503 & 563 & 448$\times$448 & 383 & 432 & 448$\times$448 & 512 & 562 \\
\cond{d} native crop, oracle & 448$\times$448 & 503 & 563 & 448$\times$448 & 383 & 432 & 448$\times$448 & 512 & 562 \\
\bottomrule
\end{tabular}
}
\end{table*}

\section{Cohort definition and the residual regime}
\label{sec:supp:cohort}
\paragraph{Three cohorts} Our primary cohort requires (i) a wrong first-pass answer, (ii) a fix by the oracle crop \cond{d} under the robust criterion (greedy answer correct and gold log-odds positive), and (iii) no such fix by the sham of \cond{d}. An alternative, \emph{confidence-filtered} cohort requires only that the first pass fail the robust criterion and that \cond{d} pass it. It admits first passes that answered correctly with non-positive gold log-odds (5--21\% of its members, \cref{tab:r4}) and rescues that a random native crop produces as well (4--11\%). A \emph{greedy-only} cohort judges the first pass, every condition and every sham by greedy correctness alone. \Cref{tab:cohorts} gives the strict partition under each. Confidence-only members land almost entirely in \rfour, because a cheaper condition that answers correctly but not confidently counts as a failure under the robust criterion. Removing them lowers the pooled \rfour\ share by 7--14 points on V$^*$+ZoomBench and 4--8 points on TextVQA. The greedy-only cohort raises \rone\ through chance flips that the strict sham rule cannot fully remove. Under all three definitions \rthree\ is the largest no-acquisition regime at 17$\times$ and \rfour\ shrinks monotonically with the first-pass budget.

\begin{table*}[h]\centering\footnotesize\setlength{\tabcolsep}{5pt}
\caption{\textbf{Strict regime partition under three cohort definitions} (pooled over first passes downsampled $\geq2\times$, in percent).}
\label{tab:cohorts}
\fittable{\begin{tabular}{llrrrrrrrrrrrrrrr}
\toprule
 & & \multicolumn{5}{c}{primary (ours)} & \multicolumn{5}{c}{confidence-filtered} & \multicolumn{5}{c}{greedy-only} \\
\cmidrule(lr){3-7}\cmidrule(lr){8-12}\cmidrule(lr){13-17}
Model & Data & $n$ & \rone & \rtwo & \rthree & \rfour & $n$ & \rone & \rtwo & \rthree & \rfour & $n$ & \rone & \rtwo & \rthree & \rfour \\
\midrule
Qwen2.5-VL-7B & V$^*$+ZoomBench & 372 & 14 & 20 & 37 & 29 & 481 & 11 & 17 & 29 & 43 & 406 & 21 & 21 & 33 & 26 \\
Qwen3-VL-8B & V$^*$+ZoomBench & 397 & 19 & 27 & 27 & 26 & 474 & 19 & 24 & 24 & 33 & 402 & 20 & 27 & 27 & 27 \\
InternVL3-8B & V$^*$+ZoomBench & 424 & 23 & 20 & 34 & 22 & 510 & 23 & 18 & 29 & 30 & 465 & 25 & 19 & 34 & 22 \\
Qwen2.5-VL-7B & TextVQA & 497 & 10 & 11 & 39 & 40 & 607 & 10 & 11 & 32 & 48 & 539 & 12 & 13 & 36 & 38 \\
Qwen3-VL-8B & TextVQA & 512 & 7 & 12 & 43 & 38 & 599 & 9 & 12 & 38 & 42 & 537 & 9 & 12 & 43 & 37 \\
\bottomrule
\end{tabular}
}
\end{table*}

\paragraph{The residual regime} \rfour\ is defined by the absence of a demonstrated cheaper fix. \Cref{tab:r4} splits it: in 94\% of its members pooled over V$^*$+ZoomBench (80--100\% per cell) every cheaper condition fails outright on the target region under the robust criterion. In the remaining 6\% a cheaper condition fixes the sample but so does its sham, so the fix cannot be attributed to the target. We count the latter as \rfour\ because the hierarchy asks for a demonstrated cheaper fix, and report them so that readers who prefer an ``unresolved'' regime can subtract them.

\begin{table*}[h]\centering\footnotesize\setlength{\tabcolsep}{5pt}
\caption{\textbf{Exclusions and the composition of \rfour.} Left: size of the confidence-filtered cohort and the members excluded from the primary cohort because the first-pass answer was correct but unconfident (conf.-only) or because a random native crop also fixed the sample (non-specific $d$). Right: size of the primary cohort, its \rfour\ members, and how many of those have every cheaper condition failing outright versus a cheaper condition succeeding together with its sham.}
\label{tab:r4}
\fittable{\begin{tabular}{llrrrrrrr}
\toprule
 & & \multicolumn{3}{c}{confidence-filtered cohort} & \multicolumn{4}{c}{primary cohort} \\
\cmidrule(lr){3-5}\cmidrule(lr){6-9}
Model & Data & $n$ & conf.-only & non-specific $d$ & $n$ & $n_{\rfour}$ & all cheaper conditions fail & cheaper condition non-specific \\
\midrule
Qwen2.5-VL-7B & V$^*$ & 141 & 18 (13\%) & 6 (4\%) & 117 & 30 & 24 (80\%) & 6 (20\%) \\
Qwen2.5-VL-7B & ZoomBench & 340 & 72 (21\%) & 13 (4\%) & 255 & 79 & 78 (99\%) & 1 (1\%) \\
Qwen2.5-VL-7B & TextVQA & 607 & 69 (11\%) & 41 (7\%) & 497 & 197 & 195 (99\%) & 2 (1\%) \\
Qwen3-VL-8B & V$^*$ & 130 & 7 (5\%) & 12 (9\%) & 111 & 15 & 13 (87\%) & 2 (13\%) \\
Qwen3-VL-8B & ZoomBench & 344 & 20 (6\%) & 38 (11\%) & 286 & 90 & 86 (96\%) & 4 (4\%) \\
Qwen3-VL-8B & TextVQA & 599 & 42 (7\%) & 45 (8\%) & 512 & 192 & 189 (98\%) & 3 (2\%) \\
InternVL3-8B & V$^*$ & 166 & 8 (5\%) & 14 (8\%) & 144 & 17 & 17 (100\%) & 0 (0\%) \\
InternVL3-8B & ZoomBench & 344 & 25 (7\%) & 39 (11\%) & 280 & 78 & 72 (92\%) & 6 (8\%) \\
\bottomrule
\end{tabular}
}
\end{table*}

\section{Strict versus lenient partition}
\label{sec:supp:lenient}
\Cref{tab:lenient} gives, per model, benchmark and budget, the strict partition used in the paper and the lenient one that drops the sham requirement. The lenient rule moves 1--10 points per cell into \rone, mostly from \rthree\ and \rfour, and the models with higher sham rates for the cached-token conditions (Qwen3-VL, InternVL3) move most.

\begin{table*}[h]\centering\footnotesize
\caption{\textbf{Strict versus lenient regime shares} (in percent, cells with at least ten rescued samples).}
\label{tab:lenient}
\fittable{\begin{tabular}{lllrrrrrrrrr}
\toprule
 & & & & \multicolumn{4}{c}{strict} & \multicolumn{4}{c}{lenient} \\
\cmidrule(lr){5-8}\cmidrule(lr){9-12}
Model & Data & First pass & $n$ & \rone & \rtwo & \rthree & \rfour & \rone & \rtwo & \rthree & \rfour \\
\midrule
InternVL3-8B & V$^*$ & 17$\times$ & 58 & 19 & 19 & 38 & 24 & 29 & 14 & 33 & 24 \\
 & V$^*$ & 4$\times$ & 44 & 30 & 25 & 39 & 7 & 34 & 23 & 36 & 7 \\
 & V$^*$ & 2.8$\times$ & 42 & 40 & 31 & 29 & 0 & 50 & 29 & 21 & 0 \\
 & ZoomBench & 17$\times$ & 108 & 13 & 10 & 33 & 44 & 19 & 12 & 31 & 38 \\
 & ZoomBench & 4$\times$ & 91 & 25 & 21 & 32 & 22 & 31 & 19 & 29 & 22 \\
 & ZoomBench & 2.8$\times$ & 81 & 26 & 23 & 37 & 14 & 33 & 23 & 30 & 14 \\
\midrule
Qwen2.5-VL-7B & TextVQA & 18$\times$ & 237 & 6 & 9 & 33 & 51 & 8 & 8 & 34 & 50 \\
 & TextVQA & 9$\times$ & 171 & 9 & 12 & 47 & 32 & 11 & 11 & 46 & 32 \\
 & TextVQA & 4$\times$ & 89 & 22 & 17 & 37 & 24 & 24 & 16 & 37 & 24 \\
 & V$^*$ & 17$\times$ & 75 & 9 & 17 & 45 & 28 & 12 & 20 & 45 & 23 \\
 & V$^*$ & 4$\times$ & 42 & 21 & 17 & 40 & 21 & 29 & 19 & 36 & 17 \\
 & ZoomBench & 17$\times$ & 128 & 9 & 14 & 40 & 38 & 12 & 15 & 37 & 37 \\
 & ZoomBench & 4$\times$ & 127 & 19 & 28 & 29 & 24 & 20 & 28 & 28 & 24 \\
\midrule
Qwen3-VL-8B & TextVQA & 17$\times$ & 239 & 4 & 10 & 38 & 48 & 5 & 9 & 38 & 47 \\
 & TextVQA & 9$\times$ & 162 & 10 & 9 & 49 & 33 & 11 & 8 & 49 & 32 \\
 & TextVQA & 4$\times$ & 111 & 11 & 23 & 45 & 22 & 14 & 20 & 45 & 22 \\
 & V$^*$ & 18$\times$ & 63 & 11 & 29 & 43 & 17 & 14 & 30 & 38 & 17 \\
 & V$^*$ & 4$\times$ & 48 & 21 & 44 & 27 & 8 & 31 & 40 & 25 & 4 \\
 & ZoomBench & 18$\times$ & 135 & 14 & 21 & 22 & 43 & 22 & 19 & 19 & 40 \\
 & ZoomBench & 4$\times$ & 151 & 26 & 27 & 26 & 21 & 28 & 28 & 23 & 21 \\
\bottomrule
\end{tabular}
}
\end{table*}

\section{Paired differences}
\Cref{tab:paired} reports, for the pairs behind the equality and sham claims in the main text, the paired difference in rescue rate with a 95\% bootstrap interval that resamples question ids (so that all budgets of a question move together), pooled over V$^*$ and ZoomBench with a first pass downsampled $\geq2\times$. The \cond{b3}$-$\cond{b2} and \cond{a3}$-$\cond{a2} intervals include zero and are narrower than five points. Every target-minus-sham interval excludes zero, including \cond{a1}'s.

\begin{table*}[h]\centering\footnotesize
\caption{\textbf{Paired differences} with question-clustered 95\% bootstrap intervals (4{,}000 resamples) and discordant counts.}
\label{tab:paired}
\fittable{\begin{tabular}{llrrrrr}
\toprule
Model & Pair & first (\%) & second (\%) & difference (pp) & 95\% CI & discordant (first only / second only) \\
\midrule
Qwen2.5-VL-7B & \cond{b3} $-$ \cond{b2} & 20.4 & 20.4 & +0.0 & [-1.4, +1.4] & 4 / 4 \\
 & \cond{a3} $-$ \cond{a2} & 14.0 & 13.2 & +0.8 & [-0.8, +2.4] & 6 / 3 \\
 & \cond{a1} $-$ sham & 4.3 & 1.3 & +3.0 & [+1.1, +4.9] & 13 / 2 \\
 & \cond{a2} $-$ sham & 13.2 & 6.7 & +6.5 & [+2.4, +10.5] & 41 / 17 \\
 & \cond{b1} $-$ sham & 23.4 & 8.6 & +14.8 & [+9.7, +19.8] & 70 / 15 \\
 & \cond{b2} $-$ sham & 20.4 & 1.6 & +18.8 & [+14.8, +23.0] & 71 / 1 \\
 & \cond{c} $-$ sham & 66.7 & 1.3 & +65.3 & [+60.2, +70.6] & 244 / 1 \\
\midrule
Qwen3-VL-8B & \cond{b3} $-$ \cond{b2} & 27.0 & 26.7 & +0.3 & [-1.7, +2.1] & 8 / 7 \\
 & \cond{a3} $-$ \cond{a2} & 20.9 & 21.2 & -0.3 & [-2.2, +1.6] & 6 / 7 \\
 & \cond{a1} $-$ sham & 8.6 & 3.5 & +5.0 & [+2.6, +7.5] & 23 / 3 \\
 & \cond{a2} $-$ sham & 21.2 & 14.4 & +6.8 & [+1.9, +11.7] & 59 / 32 \\
 & \cond{b1} $-$ sham & 37.8 & 14.9 & +22.9 & [+16.9, +28.8] & 123 / 32 \\
 & \cond{b2} $-$ sham & 26.7 & 3.8 & +22.9 & [+18.3, +27.6] & 95 / 4 \\
 & \cond{c} $-$ sham & 64.7 & 2.0 & +62.7 & [+58.1, +67.5] & 250 / 1 \\
\midrule
InternVL3-8B & \cond{b3} $-$ \cond{b2} & 22.4 & 21.2 & +1.2 & [-0.2, +2.7] & 7 / 2 \\
 & \cond{a3} $-$ \cond{a2} & 27.1 & 25.2 & +1.9 & [-0.5, +4.3] & 16 / 8 \\
 & \cond{a1} $-$ sham & 8.5 & 1.4 & +7.1 & [+4.3, +10.1] & 33 / 3 \\
 & \cond{a2} $-$ sham & 25.2 & 14.2 & +11.1 & [+5.2, +16.6] & 77 / 30 \\
 & \cond{b1} $-$ sham & 31.6 & 15.1 & +16.5 & [+10.8, +22.2] & 110 / 40 \\
 & \cond{b2} $-$ sham & 21.2 & 1.7 & +19.6 & [+15.5, +24.0] & 85 / 2 \\
 & \cond{c} $-$ sham & 73.6 & 1.7 & +71.9 & [+66.9, +76.8] & 307 / 2 \\
\bottomrule
\end{tabular}
}
\end{table*}

\section{Budget versus target coverage}
\Cref{tab:logit} reports logistic regressions of the strict \rfour\ and \rthree\ labels on $\log_2$ first-pass downsampling and $\log_2$ target coverage (ground-truth box area in first-pass tokens), with benchmark and model fixed effects and standard errors clustered by question. The budget coefficient for \rfour\ does not rest on pooling: it is positive and significant when each benchmark is fitted alone (last three rows).

\begin{table*}[h]\centering\footnotesize
\caption{\textbf{Does the budget predict the regime beyond target size?} Logit coefficients (cluster-robust standard errors), with $^{*}p<0.05$, $^{**}p<0.01$, $^{***}p<0.001$. Model fixed effects are included in every fit and benchmark fixed effects in the pooled ones. The last three rows fit each benchmark alone. Coefficients are log-odds per doubling: $+0.39$ and $+0.48$ are odds ratios of 1.5 and 1.6.}
\label{tab:logit}
\fittable{\begin{tabular}{lllrrrr}
\toprule
Data & Outcome & Predictors & $n$ & $\log_2$ downsampling & $\log_2$ target coverage & pseudo-$R^2$ \\
\midrule
V$^*$+ZoomBench & \rfour\ acquisition & both & 1193 & +0.48 (0.08)$^{***}$ & -0.02 (0.04) & 0.068 \\
V$^*$+ZoomBench & \rfour\ acquisition & downsampling only & 1193 & +0.50 (0.06)$^{***}$ & -- & 0.068 \\
V$^*$+ZoomBench & \rfour\ acquisition & coverage only & 1193 & -- & -0.11 (0.03)$^{***}$ & 0.034 \\
V$^*$+ZoomBench & \rthree\ re-tokenization & both & 1193 & -0.01 (0.06) & -0.09 (0.03)$^{**}$ & 0.019 \\
V$^*$+ZoomBench & \rthree\ re-tokenization & downsampling only & 1193 & +0.08 (0.06) & -- & 0.011 \\
V$^*$+ZoomBench & \rthree\ re-tokenization & coverage only & 1193 & -- & -0.09 (0.03)$^{***}$ & 0.019 \\
all three & \rfour\ acquisition & both & 2202 & +0.39 (0.06)$^{***}$ & -0.13 (0.03)$^{***}$ & 0.074 \\
all three & \rfour\ acquisition & downsampling only & 2202 & +0.51 (0.05)$^{***}$ & -- & 0.065 \\
all three & \rfour\ acquisition & coverage only & 2202 & -- & -0.20 (0.03)$^{***}$ & 0.055 \\
all three & \rthree\ re-tokenization & both & 2202 & -0.03 (0.05) & -0.03 (0.02) & 0.008 \\
all three & \rthree\ re-tokenization & downsampling only & 2202 & 0.00 (0.05) & -- & 0.008 \\
all three & \rthree\ re-tokenization & coverage only & 2202 & -- & -0.03 (0.02) & 0.008 \\
\midrule
V$^*$ alone & \rfour\ acquisition & both & 372 & +0.52 (0.19)$^{**}$ & 0.00 (0.10) & 0.067 \\
ZoomBench alone & \rfour\ acquisition & both & 821 & +0.46 (0.09)$^{***}$ & -0.02 (0.04) & 0.051 \\
TextVQA alone & \rfour\ acquisition & both & 1009 & +0.28 (0.11)$^{*}$ & -0.34 (0.06)$^{***}$ & 0.080 \\
\bottomrule
\end{tabular}
}
\end{table*}

\section{Budget effect with crop budget and cohort held fixed}
\label{sec:supp:budget}
Changing the first-pass budget $B$ also changes the crop budget $K=\min(B,\max(256,|R|))$, the token-grid snapping of the region, and which questions enter the cohort. Two analyses of the existing outputs separate these. First, at $B=0.2$\,MP every rescued V$^*$/ZoomBench sample has $K=256$ tokens, and 58--67\% of the rescues at the finer budgets do as well. \Cref{tab:budgetk} restricts all budgets to $K=256$, so the second look has the same pixel budget throughout, and the \rfour\ share still falls from 23\% to 0\% on V$^*$ and from 41\% to 15\% on ZoomBench. Adding $\log_2 K$ and the region's token-grid size as covariates to the regression of \cref{tab:logit} leaves the downsampling coefficient unchanged ($0.47\pm0.08$, neither covariate significant), and fitting on the $K=256$ subset alone gives $0.52\pm0.09$. Second, \cref{tab:budgetcommon} follows the questions that are zoom-rescued at both 17$\times$ and 4$\times$ by the same model: within this fixed cohort the \rfour\ share drops from 45\% to 16\%, 93 questions move from \rfour\ to a cheaper regime and 5 move the other way. The budget effect is therefore not produced by the changing crop budget or by cohort turnover. What these analyses do not separate is the first-pass information loss from the first-pass token grid itself, which change together by construction.

\begin{table*}[h]\centering\footnotesize
\caption{\textbf{Strict regime shares with the crop budget fixed at $K=256$ tokens} (V$^*$ and ZoomBench, pooled over models, in percent).}
\label{tab:budgetk}
\fittable{\begin{tabular}{llrrrrr}
\toprule
Data & First pass & $n$ & \rone & \rtwo & \rthree & \rfour \\
\midrule
V$^*$ & 17$\times$ & 196 & 13 & 21 & 42 & 23 \\
V$^*$ & 4$\times$ & 124 & 22 & 30 & 37 & 11 \\
V$^*$ & 2.8$\times$ & 37 & 41 & 32 & 27 & 0 \\
ZoomBench & 17$\times$ & 371 & 12 & 15 & 32 & 41 \\
ZoomBench & 4$\times$ & 168 & 22 & 25 & 32 & 21 \\
ZoomBench & 2.8$\times$ & 46 & 22 & 24 & 39 & 15 \\
\bottomrule
\end{tabular}
}
\end{table*}

\begin{table*}[h]\centering\footnotesize
\caption{\textbf{Questions rescued at both the 17$\times$ and the 4$\times$ first pass} by the same model: regime shares at each budget and the number of questions whose strict label moves from \rfour\ to a cheaper regime or the reverse.}
\label{tab:budgetcommon}
\fittable{\begin{tabular}{llrrrrrrrr}
\toprule
 & & & & \multicolumn{2}{c}{\rfour\ share (\%)} & \multicolumn{2}{c}{\rone\ share (\%)} & \multicolumn{2}{c}{label moves} \\
\cmidrule(lr){5-6}\cmidrule(lr){7-8}\cmidrule(lr){9-10}
Model & Data & rescued at 17$\times$ / 4$\times$ & both & at 17$\times$ & at 4$\times$ & at 17$\times$ & at 4$\times$ & \rfour$\to$cheaper & cheaper$\to$\rfour \\
\midrule
InternVL3-8B & V$^*$ & 58 / 44 & 35 & 40 & 9 & 11 & 20 & 12 & 1 \\
InternVL3-8B & ZoomBench & 108 / 91 & 56 & 48 & 25 & 9 & 27 & 15 & 2 \\
Qwen2.5-VL-7B & V$^*$ & 75 / 42 & 32 & 41 & 16 & 6 & 25 & 8 & 0 \\
Qwen2.5-VL-7B & ZoomBench & 128 / 127 & 62 & 50 & 19 & 5 & 11 & 19 & 0 \\
Qwen3-VL-8B & V$^*$ & 63 / 48 & 35 & 29 & 11 & 9 & 14 & 7 & 1 \\
Qwen3-VL-8B & ZoomBench & 135 / 151 & 84 & 50 & 13 & 13 & 27 & 32 & 1 \\
\midrule
pooled & & 567 / 503 & 304 & 45 & 16 & 9 & 21 & 93 & 5 \\
\bottomrule
\end{tabular}
}
\end{table*}

\section{Magnification sweep}
\label{sec:supp:mag}
For every zoom-rescued failure at the coarsest first pass (V$^*$ and ZoomBench at $B=0.2$\,MP, TextVQA at 0.05\,MP for the two Qwen models) we render the target region at linear magnification $m\in\{1,1.5,2,3,4,6,8,12,16,24,32\}$ of its first-pass size: the token grid of the region is scaled to $m\,g_w\times m\,g_h$ cells, rounded (at least $2\times2$, at most 1{,}024 tokens for $m\leq8$ and 4{,}096 for $m\geq12$), the seen crop (\cond{c}$_m$), the native crop (\cond{d}$_m$) and the seen crop of the random region (\cond{c}$_m$-rand) are resized to that size with the processor's own resize, and the result is appended as the second image exactly as in \cond{b2} (for InternVL3, pasted into a mean-color canvas of whole tiles). $m=1$ reproduces \cond{b2}, \cond{b3} and their shams. A rendering above the token cap is skipped, so fewer samples remain at large $m$. The curves of \cref{fig:prescription} and the per-magnification columns of \cref{tab:mag} therefore use one fixed sample, the failures whose region spans at most 16 tokens of the first-pass grid and so fits every $m\leq16$ (55--59\% of the rescued failures on ZoomBench, 89--95\% elsewhere). \Cref{tab:mag} also lists the smallest magnification at which the seen-pixel curve reaches 90\% of its maximum ($m_{90}$) and, over all rescued failures and every magnification run for them, the share fixed at any magnification (strict: at some $m$ where the random region is not fixed). Adding the failures that a cheaper condition fixes, 68--94\% of the rescued failures need no new pixels and 6--32\% do. On the smaller sample whose region fits $32\times$ (18--108 failures per model and benchmark), the seen-pixel rate changes from $8\times$ to $32\times$ by $-11$ to $+13$ points on objects (down for Qwen2.5-VL, up for InternVL3) and by $-16$ to $-18$ points on text, while native pixels stay at 92--100\%.

\begin{table*}[h]\centering\footnotesize\setlength{\tabcolsep}{3.2pt}
\caption{\textbf{Magnification sweep}: percentage of zoom-rescued failures at the coarsest first pass that each second look fixes, by magnification of the region relative to its first-pass size, on the fixed sample of failures whose region fits every $m\leq16$ (first $n$). The last three columns use all rescued failures (second $n$) and every magnification run for them, up to $32\times$.}
\label{tab:mag}
\fittable{\begin{tabular}{llrrrrrrrrrrrrrrr}
\toprule
 & & & & \multicolumn{9}{c}{magnification $m$ of the region (fixed sample)} & & \multicolumn{3}{c}{any $m$ (all rescued failures)} \\
\cmidrule(lr){5-13} \cmidrule(lr){15-17}
Model & Data & condition & $n$ & 1$\times$ & 1.5$\times$ & 2$\times$ & 3$\times$ & 4$\times$ & 6$\times$ & 8$\times$ & 12$\times$ & 16$\times$ & $m_{90}$ & $n$ & fixed & strict \\
\midrule
Qwen2.5-VL-7B & V$^*$ & seen px, re-tokenized & 68 & 13 & 38 & 54 & 56 & 62 & 74 & 72 & 71 & 69 & 6$\times$ & 75 & 88 & 88 \\
 &  & native px, same size &  & 13 & 37 & 51 & 74 & 90 & 94 & 100 & 100 & 97 & -- & & 97 & 97 \\
 &  & seen px, random region &  & 1 & 0 & 0 & 0 & 0 & 3 & 4 & 1 & 4 & -- & & 9 & -- \\
\addlinespace[2pt]
Qwen2.5-VL-7B & ZoomBench & seen px, re-tokenized & 70 & 9 & 19 & 17 & 37 & 43 & 50 & 57 & 56 & 57 & 8$\times$ & 128 & 77 & 77 \\
 &  & native px, same size &  & 10 & 20 & 27 & 49 & 67 & 84 & 90 & 90 & 87 & -- & & 97 & 96 \\
 &  & seen px, random region &  & 0 & 0 & 0 & 0 & 1 & 0 & 1 & 1 & 3 & -- & & 5 & -- \\
\addlinespace[2pt]
Qwen2.5-VL-7B & TextVQA & seen px, re-tokenized & 211 & 5 & 28 & 36 & 44 & 49 & 47 & 47 & 41 & 38 & 4$\times$ & 237 & 65 & 65 \\
 &  & native px, same size &  & 6 & 41 & 71 & 91 & 97 & 94 & 95 & 96 & 94 & -- & & 100 & 100 \\
 &  & seen px, random region &  & 0 & 0 & 0 & 0 & 0 & 0 & 0 & 0 & 0 & -- & & 2 & -- \\
\addlinespace[2pt]
Qwen3-VL-8B & V$^*$ & seen px, re-tokenized & 60 & 22 & 30 & 37 & 52 & 62 & 75 & 82 & 85 & 82 & 8$\times$ & 63 & 94 & 94 \\
 &  & native px, same size &  & 22 & 42 & 50 & 83 & 92 & 97 & 100 & 100 & 100 & -- & & 100 & 100 \\
 &  & seen px, random region &  & 8 & 10 & 7 & 3 & 3 & 5 & 3 & 2 & 2 & -- & & 11 & -- \\
\addlinespace[2pt]
Qwen3-VL-8B & ZoomBench & seen px, re-tokenized & 76 & 12 & 13 & 17 & 34 & 39 & 50 & 54 & 63 & 62 & 12$\times$ & 135 & 73 & 70 \\
 &  & native px, same size &  & 12 & 22 & 34 & 59 & 80 & 92 & 96 & 97 & 99 & -- & & 98 & 97 \\
 &  & seen px, random region &  & 3 & 3 & 1 & 3 & 1 & 1 & 3 & 3 & 3 & -- & & 7 & -- \\
\addlinespace[2pt]
Qwen3-VL-8B & TextVQA & seen px, re-tokenized & 224 & 4 & 23 & 39 & 52 & 56 & 61 & 58 & 55 & 52 & 4$\times$ & 239 & 74 & 73 \\
 &  & native px, same size &  & 4 & 32 & 68 & 90 & 90 & 93 & 94 & 95 & 96 & -- & & 100 & 99 \\
 &  & seen px, random region &  & 2 & 2 & 2 & 3 & 4 & 3 & 3 & 3 & 4 & -- & & 6 & -- \\
\addlinespace[2pt]
InternVL3-8B & V$^*$ & seen px, re-tokenized & 55 & 7 & 13 & 24 & 38 & 49 & 56 & 69 & 71 & 75 & 8$\times$ & 58 & 84 & 83 \\
 &  & native px, same size &  & 7 & 25 & 45 & 80 & 87 & 96 & 100 & 98 & 98 & -- & & 100 & 100 \\
 &  & seen px, random region &  & 2 & 0 & 0 & 2 & 0 & 4 & 4 & 5 & 5 & -- & & 14 & -- \\
\addlinespace[2pt]
InternVL3-8B & ZoomBench & seen px, re-tokenized & 64 & 6 & 12 & 16 & 25 & 39 & 45 & 50 & 55 & 59 & 12$\times$ & 108 & 69 & 69 \\
 &  & native px, same size &  & 6 & 16 & 33 & 50 & 61 & 75 & 88 & 88 & 91 & -- & & 95 & 95 \\
 &  & seen px, random region &  & 2 & 2 & 2 & 2 & 2 & 3 & 3 & 8 & 5 & -- & & 8 & -- \\
\addlinespace[2pt]
\bottomrule
\end{tabular}
}
\end{table*}

\section{Accuracy and cohort sizes per model, benchmark and budget}
\Cref{tab:stage1full} lists, for every cell, the robust accuracy of the first pass and of the oracle zoom, the number of zoom-rescued and zoom-hurt questions and, where the first pass is downsampled at least 2$\times$, the strict regime shares. The 2{,}753 zoom-rescued failures of the paper are the sum of its rescued column, and the partition uses the 2{,}202 of them in the downsampled cells.
\begin{table*}[h]\centering\footnotesize
\caption{\textbf{Accuracy, cohort sizes and regime shares per model, benchmark and first-pass budget.} Robust accuracy without and with the oracle zoom, the number of zoom-rescued and zoom-hurt samples, and the strict regime shares of the rescued set where the first pass was downsampled $\geq2\times$.}
\label{tab:stage1full}
\fittable{\begin{tabular}{llrrrrrrl}
\toprule
Model & Benchmark & First pass & $n$ & Base & +Zoom & Rescued & Hurt & R1 / R2 / R3 / R4 (\%) \\
\midrule
Qwen2.5-VL-7B & TextVQA & 18$\times$ & 1000 & 0.439 & 0.701 & 237 & 29 & 6 / 9 / 33 / 51 \\
 &  & 9$\times$ & 1000 & 0.565 & 0.748 & 171 & 20 & 9 / 12 / 47 / 32 \\
 &  & 4$\times$ & 1000 & 0.685 & 0.773 & 89 & 25 & 22 / 17 / 37 / 24 \\
 &  & native & 1000 & 0.764 & 0.786 & 21 & 30 & -- \\
Qwen2.5-VL-7B & V$^*$ & 17$\times$ & 191 & 0.476 & 0.921 & 75 & 6 & 9 / 17 / 45 / 28 \\
 &  & 4$\times$ & 191 & 0.696 & 0.932 & 42 & 5 & 21 / 17 / 40 / 21 \\
 &  & 1.1$\times$ & 191 & 0.775 & 0.948 & 31 & 5 & -- \\
 &  & native & 191 & 0.791 & 0.937 & 29 & 4 & -- \\
Qwen2.5-VL-7B & ZoomBench & 17$\times$ & 844 & 0.222 & 0.389 & 128 & 33 & 9 / 14 / 40 / 38 \\
 &  & 4$\times$ & 844 & 0.293 & 0.444 & 127 & 38 & 19 / 28 / 29 / 24 \\
 &  & 1.1$\times$ & 844 & 0.370 & 0.498 & 93 & 29 & -- \\
 &  & native & 844 & 0.380 & 0.476 & 82 & 31 & -- \\
\midrule
Qwen3-VL-8B & TextVQA & 17$\times$ & 1000 & 0.441 & 0.687 & 239 & 24 & 4 / 10 / 38 / 48 \\
 &  & 9$\times$ & 1000 & 0.573 & 0.734 & 162 & 33 & 10 / 9 / 49 / 33 \\
 &  & 4$\times$ & 1000 & 0.665 & 0.756 & 111 & 44 & 11 / 23 / 45 / 22 \\
 &  & native & 1000 & 0.753 & 0.752 & 19 & 39 & -- \\
Qwen3-VL-8B & V$^*$ & 18$\times$ & 191 & 0.550 & 0.916 & 63 & 5 & 11 / 29 / 43 / 17 \\
 &  & 4$\times$ & 191 & 0.691 & 0.942 & 48 & 7 & 21 / 44 / 27 / 8 \\
 &  & 1.1$\times$ & 191 & 0.838 & 0.948 & 23 & 6 & -- \\
 &  & native & 191 & 0.843 & 0.953 & 21 & 4 & -- \\
Qwen3-VL-8B & ZoomBench & 18$\times$ & 844 & 0.335 & 0.470 & 135 & 53 & 14 / 21 / 22 / 43 \\
 &  & 4$\times$ & 844 & 0.378 & 0.539 & 151 & 41 & 26 / 27 / 26 / 21 \\
 &  & 1.1$\times$ & 844 & 0.436 & 0.558 & 117 & 38 & -- \\
 &  & native & 844 & 0.428 & 0.538 & 115 & 39 & -- \\
\midrule
InternVL3-8B & V$^*$ & 17$\times$ & 191 & 0.586 & 0.906 & 58 & 5 & 19 / 19 / 38 / 24 \\
 &  & 4$\times$ & 191 & 0.696 & 0.953 & 44 & 3 & 30 / 25 / 39 / 7 \\
 &  & 2.8$\times$ & 191 & 0.696 & 0.932 & 42 & 3 & 40 / 31 / 29 / 0 \\
InternVL3-8B & ZoomBench & 17$\times$ & 844 & 0.352 & 0.451 & 108 & 43 & 13 / 10 / 33 / 44 \\
 &  & 4$\times$ & 844 & 0.368 & 0.454 & 91 & 43 & 25 / 21 / 32 / 22 \\
 &  & 2.8$\times$ & 844 & 0.405 & 0.477 & 81 & 41 & 26 / 23 / 37 / 14 \\
\bottomrule
\end{tabular}
}
\end{table*}

\section{Self-grounding: does the model know where to look?}
\label{sec:supp:ground}
The main experiments use ground-truth regions. To measure how far that oracle is from what the models can do themselves, we asked each model, on its own first-pass image, for the region needed to answer the question. Qwen2.5-VL receives one question-conditioned prompt and returns boxes in input-pixel coordinates, and Qwen3-VL and InternVL3 return 0--1000 normalized coordinates. InternVL3 uses its native referring-grounding format, with the referred phrase first produced from the question text alone, because its free-form boxes were unusable. We then reran every condition with the model's own box in place of the ground-truth box on the zoom-rescued cohort, with the same padding, snapping, crop budget and sham construction. \Cref{tab:ground} summarizes the results on the rescued failures. Over all questions, at 17$\times$ the models' boxes cover the center of the target region for 19--62\% (IoU $\geq0.5$: 1--38\%). Qwen2.5-VL declines to give a box for 38\% of V$^*$ questions, Qwen3-VL localizes best, and InternVL3 returns large boxes (median 6--8\% of the image) that rarely fit the target. Localization improves with the first-pass budget at about the rate recognition does (Qwen2.5-VL on V$^*$: 19\% to 46\% of centers covered from 17$\times$ to 4$\times$), and within the rescued cohort no regime is localized markedly better than the others: near-sighted (\rthree) failures have their target covered 28--53\% of the time at 17$\times$, like the rest. With the model's own region, 28--84\% of the rescued failures (33--78\% at 17$\times$) are fixed by no condition at all, not even the native crop of that region, and the strict label agrees with the ground-truth one for 9--55\% of samples. The disagreement is almost entirely localization: pooled over models, when the own box covers the target the partition is 11/16/24/14\% for \rone--\rfour\ with 34\% unfixable, against 7/8/13/9\% with 62\% unfixable when it does not. The regimes therefore describe what recognition needs once the region is known. At coarse first passes, knowing the region is the larger obstacle for these models.

\begin{table*}[h]\centering\footnotesize\setlength{\tabcolsep}{5pt}
\caption{\textbf{Self-grounding.} Share of zoom-rescued failures whose own box covers the target center, the strict partition with ground-truth regions and with the model's own region (``none'': no condition fixes the sample, including the native crop of the own region), and the share of samples with the same label under both. $n$ counts the rescued failures for which the model produced a box at all. Qwen2.5-VL produced none for 35 of 75 V$^*$ and 16 of 128 ZoomBench failures at 17$\times$, which would add to ``none''. Percentages.}
\label{tab:ground}
\fittable{\begin{tabular}{llrrrrrrrrrrrrr}
\toprule
 & & & \multicolumn{2}{c}{own box} & \multicolumn{4}{c}{GT region} & \multicolumn{5}{c}{model's own region} & \\
\cmidrule(lr){4-5}\cmidrule(lr){6-9}\cmidrule(lr){10-14}
Model & Data & First pass & covers target & $n$ & \rone & \rtwo & \rthree & \rfour & \rone & \rtwo & \rthree & \rfour & none & same label \\
\midrule
InternVL3-8B & V$^*$ & 17$\times$ & 29\% & 58 & 19 & 19 & 38 & 24 & 2 & 2 & 14 & 5 & 78 & 12\% \\
InternVL3-8B & V$^*$ & 4$\times$ & 41\% & 44 & 30 & 25 & 39 & 7 & 7 & 0 & 7 & 2 & 84 & 9\% \\
InternVL3-8B & ZoomBench & 17$\times$ & 45\% & 108 & 13 & 10 & 33 & 44 & 5 & 1 & 12 & 11 & 71 & 20\% \\
InternVL3-8B & ZoomBench & 4$\times$ & 66\% & 91 & 25 & 21 & 32 & 22 & 7 & 7 & 4 & 5 & 77 & 15\% \\
Qwen2.5-VL-7B & V$^*$ & 17$\times$ & 12\% & 40 & 12 & 25 & 48 & 15 & 0 & 18 & 40 & 8 & 35 & 45\% \\
Qwen2.5-VL-7B & V$^*$ & 4$\times$ & 17\% & 23 & 22 & 22 & 35 & 22 & 17 & 13 & 30 & 4 & 35 & 52\% \\
Qwen2.5-VL-7B & ZoomBench & 17$\times$ & 30\% & 112 & 8 & 12 & 41 & 39 & 4 & 7 & 18 & 16 & 55 & 29\% \\
Qwen2.5-VL-7B & ZoomBench & 4$\times$ & 50\% & 126 & 19 & 28 & 29 & 25 & 13 & 18 & 22 & 11 & 35 & 48\% \\
Qwen3-VL-8B & V$^*$ & 17$\times$ & 33\% & 63 & 11 & 29 & 43 & 17 & 3 & 14 & 16 & 22 & 44 & 40\% \\
Qwen3-VL-8B & V$^*$ & 4$\times$ & 46\% & 48 & 21 & 44 & 27 & 8 & 15 & 23 & 19 & 4 & 40 & 48\% \\
Qwen3-VL-8B & ZoomBench & 17$\times$ & 57\% & 135 & 14 & 21 & 22 & 43 & 10 & 17 & 19 & 20 & 33 & 55\% \\
Qwen3-VL-8B & ZoomBench & 4$\times$ & 56\% & 151 & 26 & 27 & 26 & 21 & 18 & 19 & 25 & 10 & 28 & 53\% \\
\bottomrule
\end{tabular}
}
\end{table*}

\section{Robustness of the partition}
\label{sec:supp:robust}
\begin{figure*}[h]\centering
\includegraphics[width=\linewidth]{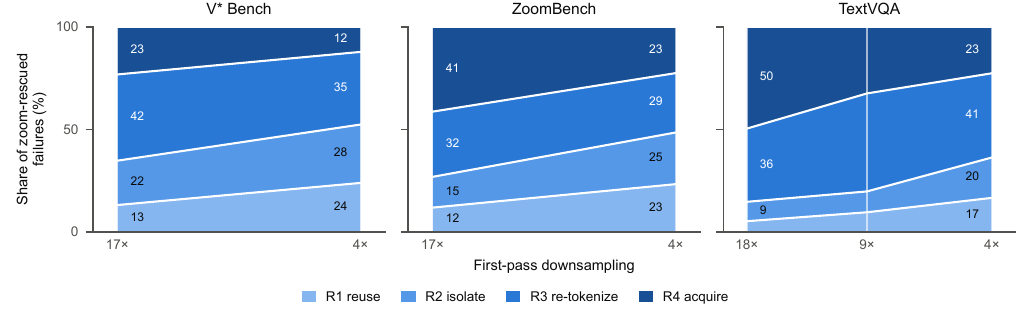}
\caption{\textbf{Regimes against the first-pass budget.} Strict regime shares of zoom-rescued failures, averaged over models, as stacked bands over the first-pass downsampling factor (coarse on the left). Numbers are percentages at the measured budgets ($n$ = 196/134 V$^*$, 371/369 ZoomBench, 476/333/200 TextVQA). Per-model changes are in \cref{fig:dumbbell}.}
\label{fig:nativecurve}
\end{figure*}

\begin{figure*}[h]\centering
\includegraphics[width=\linewidth]{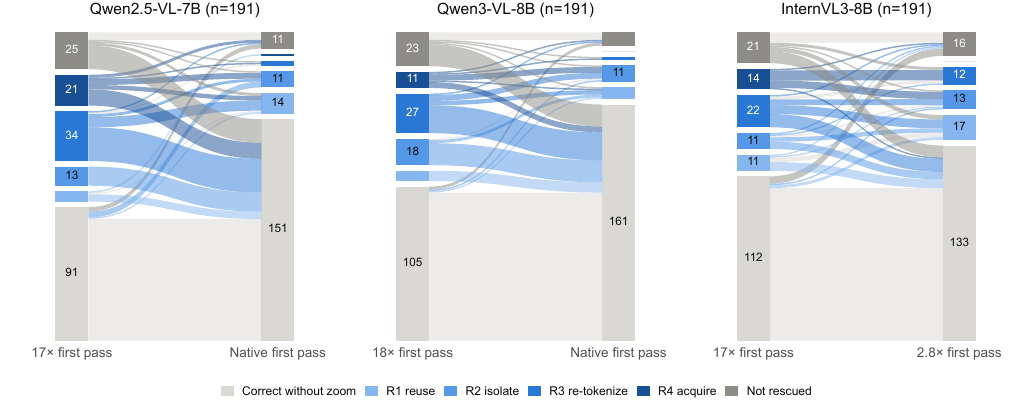}
\caption{\textbf{From the coarsest to the finest first pass.} Each V$^*$ question's outcome at the coarsest budget (left) and at the finest one (right: native resolution, 2.8$\times$ for InternVL3), per model. Flows are colored by regime, and same-regime flows are gray. ``Not rescued'' covers questions the oracle zoom does not fix robustly or fixes no better than a random crop.}
\label{fig:flow}
\end{figure*}

A zoom with twice the context padding still fixes 68--71\% of rescues (91--94\% on V$^*$ and less on ZoomBench, where the padded crop often covers most of the image), and 98--100\% of the underlying failures persist under a paraphrased instruction (\cref{tab:robust}). These checks test the cohort, not the per-sample labels. The resource hierarchy is violated (a cheaper regime fixes a sample while upsampling does not) for 0--13\% of the samples of a cell (\cref{fig:ladder}). Dropping the sham requirement changes the label of 7.0\% of the rescued V$^*$ and ZoomBench failures and of 1.6\% of the TextVQA ones, and moves 1--10 points per cell into \rone\ (\cref{tab:lenient}). A greedy-only cohort moves no share by more than 7 points, and a confidence-filtered cohort that admits correct-but-unconfident first passes inflates Qwen2.5-VL's \rfour\ share from 29\% to 43\% (Appendix~\ref{sec:supp:cohort}). The ordering of the regimes is the same throughout. \Cref{fig:nativecurve} shows the partition averaged over models as a function of the budget, and \cref{fig:flow} follows every V$^*$ question from the coarsest to the finest first pass: of the questions zoom-rescued at 17$\times$, 69--71\% (Qwen2.5-VL, Qwen3-VL) are correct without any zoom once the model sees the native image, 36\% for InternVL3 whose finest first pass is 2.8$\times$. Of those that remain zoom-rescued, at most one per model still needs new pixels.

\paragraph{A second sham placement} Every label above rests on one random region per sample and budget. We drew a second, independent region for every sample (Appendix~\ref{sec:supp:impl}), reran the eight shams on it and labeled every failure again (\cref{tab:sham2}). Before each new draw the rerun recomputed stored shams of the first placement (64 rows in each of the 20 cells) and reproduced every answer and every log-odds exactly. Cohort membership depends on the placement as well, because a failure counts as rescued only if the sham of \cond{d} does not fix it: 1{,}165 of the 1{,}193 V$^*$ and ZoomBench failures and 993 of the 1{,}009 TextVQA failures are rescued under both placements. Of these, 95.4\% and 99.1\% keep their regime label. The two draws return the same region for 8\% and 11\% of them, because their target regions cover a large part of the token grid (a median of 43\% and 30\%, against 3\% and 10\% for the others) and leave few positions. On the others the agreement is 95.0\% and 99.0\%. The changed labels have no preferred direction: on the object benchmarks 27 failures move to a cheaper regime and 27 to a more expensive one, and on TextVQA 3 and 6. Pooled over budgets, no regime share of any model moves by more than 1.8 points. Shares of single cells move by up to 1.5 points on TextVQA, 2.8 on ZoomBench and 6.6 on V$^*$, whose cells hold 42--75 failures. Under the second placement all four regimes are populated in every cell, including the one that is empty in the main text (InternVL3 on V$^*$ at 2.8$\times$, where one of 42 failures now needs new pixels). Averaged over models, \rfour\ holds 23\%, 40\% and 49\% of the rescues on V$^*$, ZoomBench and TextVQA at 17$\times$ (23\%, 41\% and 50\% under the first placement) and 10\%, 24\% and 23\% at 4$\times$ (12\%, 23\% and 23\%). At 17$\times$ \cond{c} fixes 46--71\% of the rescued failures and its sham 0--3\% (47--73\% and 1--3\%).

\begin{table*}[h]\centering\footnotesize
\caption{\textbf{Robustness} of the rescued cohort, pooled over V$^*$ and ZoomBench with a first pass $\geq2\times$ downsampled: share of rescues that a zoom with twice the padding still fixes, share of first-pass failures that persist under a paraphrased instruction, and share of samples fixed by native pixels at the first-pass density (\cond{b3}) but not by the seen pixels at that density (\cond{b2}).}
\label{tab:robust}
\fittable{\begin{tabular}{lrrrr}
\toprule
Model & $n$ & fixed with 2$\times$ padding & failure persists under paraphrase & \cond{b3} fixes, \cond{b2} does not \\
\midrule
Qwen2.5-VL-7B & 372 & 68\% & 100\% & 1\% \\
Qwen3-VL-8B & 397 & 71\% & 99\% & 2\% \\
InternVL3-8B & 424 & 68\% & 98\% & 2\% \\
\bottomrule
\end{tabular}
}
\end{table*}

\begin{table*}[h]\centering\footnotesize
\caption{\textbf{Regime labels under two independent sham placements.} Zoom-rescued failures under the first placement (the one used in the paper), under the second and under both, the share of the latter whose regime label is unchanged with Cohen's $\kappa$, and the strict regime shares under each placement (first passes downsampled $\geq2\times$, pooled over budgets).}
\label{tab:sham2}
\fittable{\begin{tabular}{@{}llrrrrrll@{}}
\toprule
 & & \multicolumn{3}{c}{Zoom-rescued failures} & Labels & & \multicolumn{2}{c}{\rone\ / \rtwo\ / \rthree\ / \rfour\ (\%)} \\
\cmidrule(lr){3-5}\cmidrule(l){8-9}
Model & Data & first & second & both & unchanged (\%) & $\kappa$ & first placement & second placement \\
\midrule
Qwen2.5-VL-7B & V$^*$+ZoomBench & 372 & 373 & 367 & 97.8 & 0.97 & 14 / 20 / 37 / 29 & 14 / 21 / 37 / 28 \\
Qwen3-VL-8B & V$^*$+ZoomBench & 397 & 394 & 386 & 93.0 & 0.91 & 19 / 27 / 27 / 26 & 20 / 25 / 28 / 27 \\
InternVL3-8B & V$^*$+ZoomBench & 424 & 419 & 412 & 95.4 & 0.94 & 23 / 20 / 34 / 22 & 24 / 18 / 35 / 23 \\
pooled & V$^*$+ZoomBench & 1{,}193 & 1{,}186 & 1{,}165 & 95.4 & 0.94 & 19 / 22 / 33 / 26 & 19 / 21 / 33 / 26 \\
\midrule
Qwen2.5-VL-7B & TextVQA & 497 & 499 & 490 & 99.2 & 0.99 & 10 / 11 / 39 / 40 & 10 / 12 / 38 / 40 \\
Qwen3-VL-8B & TextVQA & 512 & 513 & 503 & 99.0 & 0.98 & 7 / 12 / 43 / 38 & 8 / 12 / 44 / 37 \\
pooled & TextVQA & 1{,}009 & 1{,}012 & 993 & 99.1 & 0.99 & 9 / 12 / 41 / 39 & 9 / 12 / 41 / 38 \\
\bottomrule
\end{tabular}
}
\end{table*}

\begin{figure*}[h]\centering
\includegraphics[width=\linewidth]{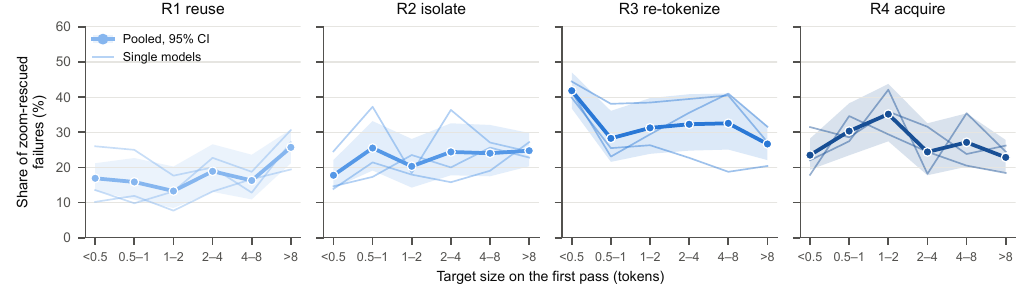}
\caption{\textbf{Regime share against target size on the first pass.} For each regime, the share of zoom-rescued V$^*$+ZoomBench failures (first pass $\geq2\times$ downsampled) that falls into it, by the number of first-pass tokens the target occupied. Bold: pooled over models with a 95\% Wilson band ($n$ per bin: 349, 145, 128, 127, 129, 315), thin: single models. The \rthree\ share is largest for sub-token targets, reuse grows with target size, and the \rfour\ share is flat in target size.}
\label{fig:phase}
\end{figure*}

\section{Zoom policy, all thresholds}
\Cref{tab:policyfull} varies the confidence threshold of the policy of \cref{tab:policy} and adds a policy that tries cached tokens before re-tokenizing.
\Cref{tab:policylong} repeats \cref{tab:policy} with the always-acquire reference and the encoder calls.
\begin{table*}[h]\centering\footnotesize
\caption{\textbf{Acquisitions, encoder calls and rescues kept by policy.} \Cref{tab:policy} with the always-acquire reference and the cells that are constant by construction, on the zoom-rescued V$^*$+ZoomBench failures at 17$\times$ when the policy stops at $p_{\max}\geq0.9$.}
\label{tab:policylong}
\fittable{\begin{tabular}{llrrr}
\toprule
Model & Policy & acquisitions (\%) & encoder calls (\%) & rescues kept (\%) \\
\midrule
Qwen2.5-VL-7B & always acquire & 100 & 100 & 100 \\
 & re-tokenize only & 0 & 100 & 63 \\
 & re-tokenize, acquire if $p_{\max}<0.9$ & 68 & 168 & 98 \\
\addlinespace[2pt]
Qwen3-VL-8B & always acquire & 100 & 100 & 100 \\
 & re-tokenize only & 0 & 100 & 55 \\
 & re-tokenize, acquire if $p_{\max}<0.9$ & 36 & 136 & 81 \\
\addlinespace[2pt]
InternVL3-8B & always acquire & 100 & 100 & 100 \\
 & re-tokenize only & 0 & 100 & 58 \\
 & re-tokenize, acquire if $p_{\max}<0.9$ & 61 & 161 & 93 \\
\addlinespace[2pt]
\bottomrule
\end{tabular}
}
\end{table*}
\begin{table*}[h]\centering\footnotesize
\caption{\textbf{The re-tokenize-first policy at four confidence thresholds, and a policy that tries cached tokens first.} Same quantities as \cref{tab:policy}. Stopping on cached-token answers loses many rescues: those answers are confidently wrong more often than answers from re-tokenized pixels.}
\label{tab:policyfull}
\fittable{\begin{tabular}{llrrr}
\toprule
Model & Policy & acquisitions (\%) & encoder calls (\%) & rescues kept (\%) \\
\midrule
Qwen2.5-VL-7B & always acquire & 100 & 100 & 100 \\
 & re-tokenize only & 0 & 100 & 63 \\
 & re-tokenize, acquire if $p_{\max}<0.7$ & 47 & 147 & 91 \\
 & re-tokenize, acquire if $p_{\max}<0.8$ & 60 & 160 & 96 \\
 & re-tokenize, acquire if $p_{\max}<0.9$ & 68 & 168 & 98 \\
 & re-tokenize, acquire if $p_{\max}<0.95$ & 80 & 180 & 99 \\
 & cached $\to$ re-tokenize $\to$ acquire ($p_{\max}<0.7$) & 37 & 116 & 77 \\
 & cached $\to$ re-tokenize $\to$ acquire ($p_{\max}<0.8$) & 52 & 141 & 87 \\
 & cached $\to$ re-tokenize $\to$ acquire ($p_{\max}<0.9$) & 66 & 162 & 95 \\
 & cached $\to$ re-tokenize $\to$ acquire ($p_{\max}<0.95$) & 79 & 177 & 98 \\
\addlinespace[2pt]
Qwen3-VL-8B & always acquire & 100 & 100 & 100 \\
 & re-tokenize only & 0 & 100 & 55 \\
 & re-tokenize, acquire if $p_{\max}<0.7$ & 17 & 117 & 69 \\
 & re-tokenize, acquire if $p_{\max}<0.8$ & 26 & 126 & 75 \\
 & re-tokenize, acquire if $p_{\max}<0.9$ & 36 & 136 & 81 \\
 & re-tokenize, acquire if $p_{\max}<0.95$ & 43 & 143 & 86 \\
 & cached $\to$ re-tokenize $\to$ acquire ($p_{\max}<0.7$) & 8 & 38 & 34 \\
 & cached $\to$ re-tokenize $\to$ acquire ($p_{\max}<0.8$) & 14 & 61 & 47 \\
 & cached $\to$ re-tokenize $\to$ acquire ($p_{\max}<0.9$) & 25 & 81 & 58 \\
 & cached $\to$ re-tokenize $\to$ acquire ($p_{\max}<0.95$) & 31 & 96 & 65 \\
\addlinespace[2pt]
InternVL3-8B & always acquire & 100 & 100 & 100 \\
 & re-tokenize only & 0 & 100 & 58 \\
 & re-tokenize, acquire if $p_{\max}<0.7$ & 35 & 135 & 83 \\
 & re-tokenize, acquire if $p_{\max}<0.8$ & 48 & 148 & 88 \\
 & re-tokenize, acquire if $p_{\max}<0.9$ & 61 & 161 & 93 \\
 & re-tokenize, acquire if $p_{\max}<0.95$ & 71 & 171 & 97 \\
 & cached $\to$ re-tokenize $\to$ acquire ($p_{\max}<0.7$) & 19 & 70 & 52 \\
 & cached $\to$ re-tokenize $\to$ acquire ($p_{\max}<0.8$) & 32 & 96 & 62 \\
 & cached $\to$ re-tokenize $\to$ acquire ($p_{\max}<0.9$) & 46 & 124 & 74 \\
 & cached $\to$ re-tokenize $\to$ acquire ($p_{\max}<0.95$) & 61 & 148 & 86 \\
\addlinespace[2pt]
\bottomrule
\end{tabular}
}
\end{table*}

\section{Zoom-hurt samples}
\label{sec:supp:hurt}
Samples that the first pass answers robustly and the oracle crop does not (\cref{tab:stage1full}) are 2--4\% of V$^*$ questions, 3--6\% of ZoomBench and 2--4\% of TextVQA per cell. They are not analyzed further here because the question of this paper is conditional on a rescue. A zoom policy would additionally have to decide \emph{whether} to look again, which is the question studied by CounterCredit~\cite{countercredit2026}.

\section{Additional results}
\begin{figure*}[h]\centering
\includegraphics[width=\linewidth]{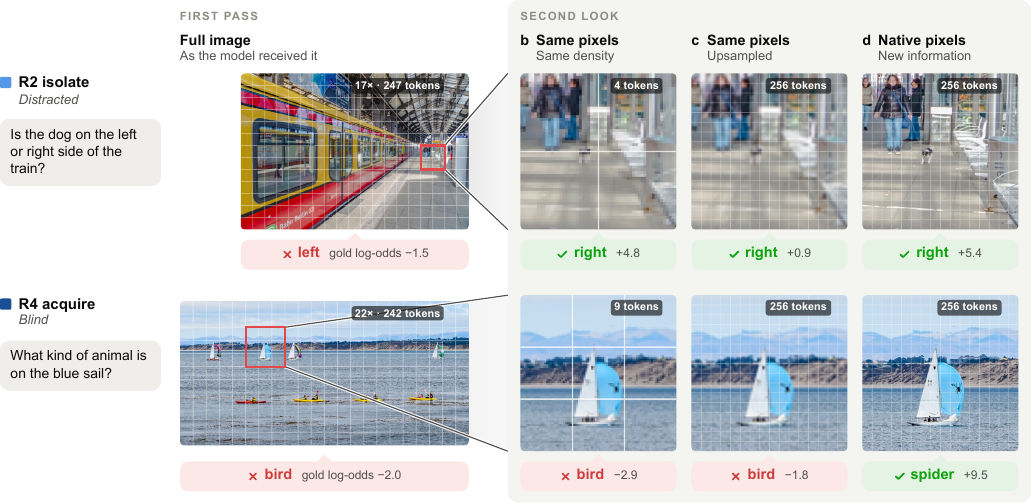}
\caption{\textbf{Two more zoom-rescued V$^*$ failures} (Qwen2.5-VL-7B, coarsest first pass, the \rthree\ example being the bag of \cref{fig:teaser}): the inputs the model actually received under \cond{b2}, \cond{c} and \cond{d} with their token grids, and its replies with the gold-answer log-odds. Isolating the region at the density the model already saw suffices for the dog (\rtwo). Only native pixels reveal the spider on the sail, which every cheaper look calls a bird (\rfour).}
\label{fig:examples}
\end{figure*}

\begin{figure*}[h]\centering
\includegraphics[width=\linewidth]{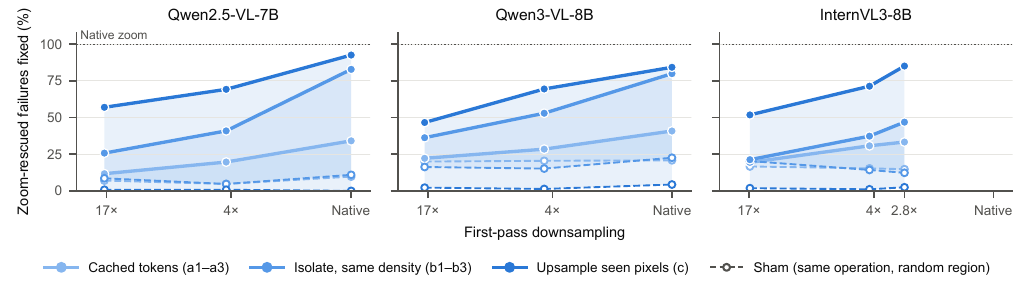}\\[4pt]
\includegraphics[width=\linewidth]{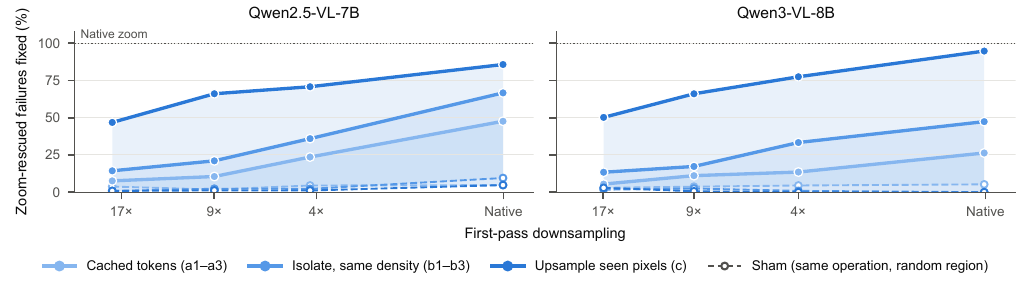}
\caption{\textbf{The sweep of \cref{fig:sweep} on ZoomBench (top) and TextVQA (bottom).} The ordering of the three families is the same as on V$^*$: upsampling the seen pixels fixes the most, cached tokens the least.}\label{fig:sweep_zb}
\end{figure*}

\begin{figure*}[h]\centering
\includegraphics[width=\linewidth]{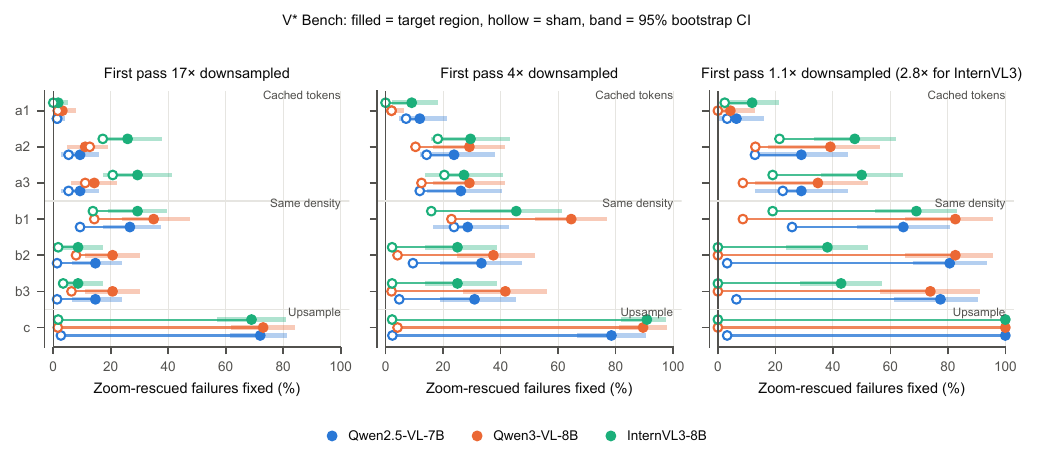}\\
\includegraphics[width=\linewidth]{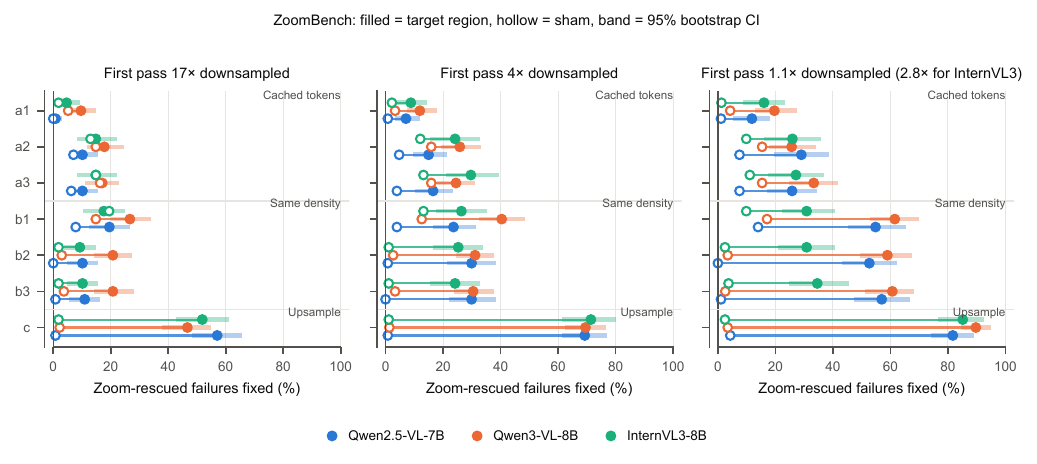}
\caption{\textbf{Every condition against its sham.} Share of zoom-rescued failures fixed by each condition on the target region (filled) and by its sham (hollow), with 95\% bootstrap intervals. Rows are the conditions in the order of \cref{tab:conditions}, columns first-pass budgets.}\label{fig:forest}
\end{figure*}

\begin{figure*}[h]\centering
\includegraphics[width=\linewidth]{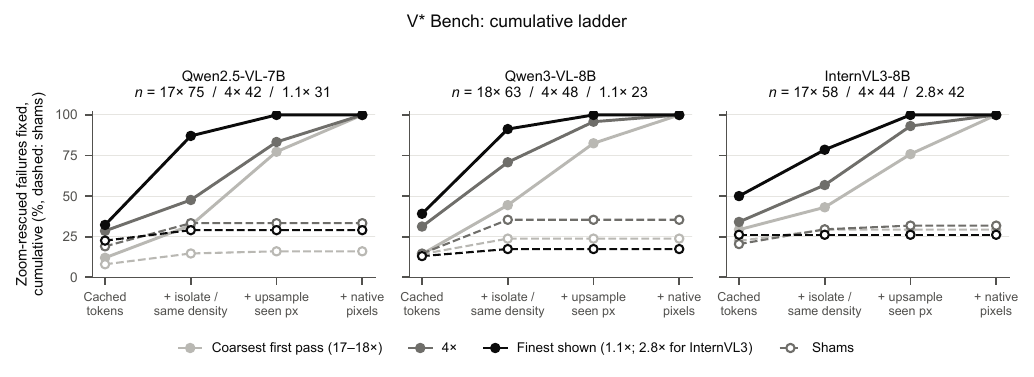}\\
\includegraphics[width=\linewidth]{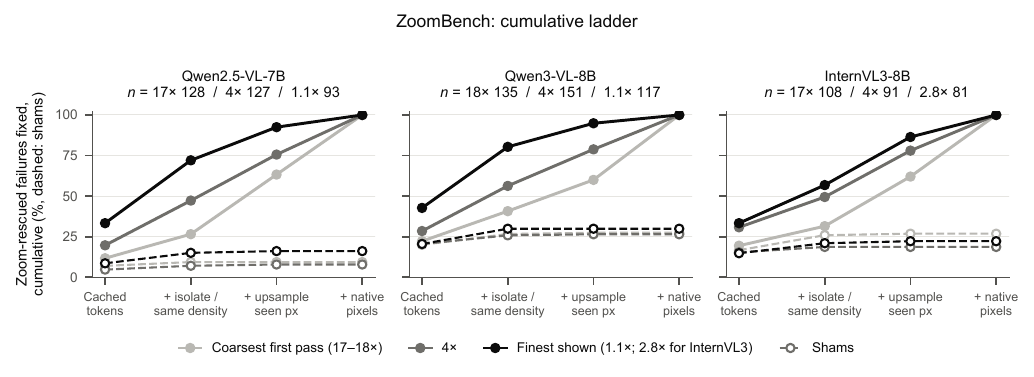}
\caption{\textbf{Cumulative ladder.} The ladder adds the conditions in order of cost. Share of zoom-rescued failures fixed by at least one condition up to each step (solid) and by at least one of their shams (dashed), per first-pass budget. Hierarchy violations (a cheaper step fixes a sample but upsampling does not) are 0--13\% per cell.}\label{fig:ladder}
\end{figure*}

\begin{figure*}[h]\centering
\includegraphics[width=\linewidth]{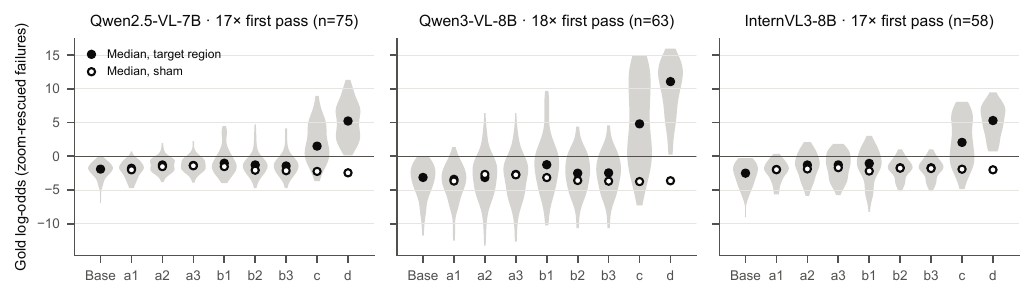}
\caption{\textbf{Graded effects.} Gold log-odds of the first pass (Base) and under each condition of \cref{tab:conditions} on zoom-rescued V$^*$ failures at the coarsest budget (medians: filled target, hollow sham).}\label{fig:graded}
\end{figure*}

\begin{figure*}[h]\centering
\includegraphics[width=\linewidth]{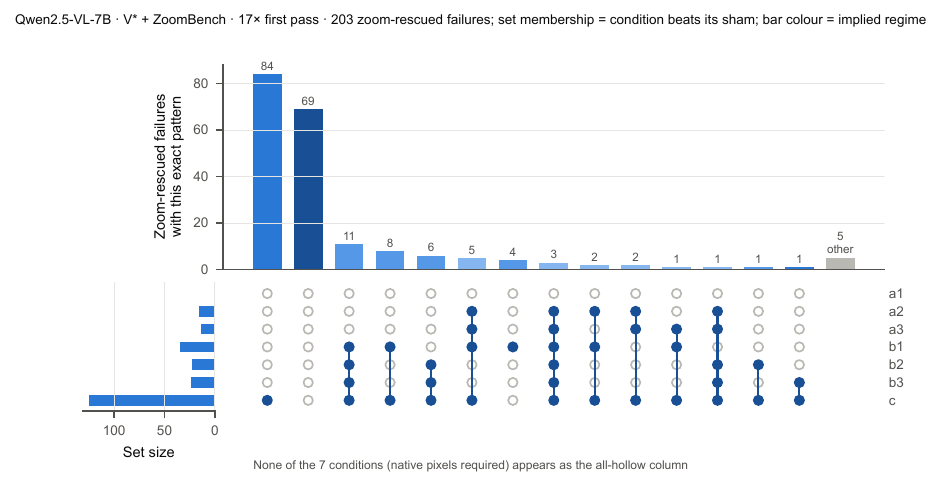}
\caption{\textbf{Which conditions fix the same failures.} UpSet view of the sets of failures that each condition fixes while its sham does not (Qwen2.5-VL, V$^*$+ZoomBench, 17$\times$): the dominant patterns are ``\cond{c} only'' and ``none'' (new pixels required), then the \cond{b} conditions with \cond{c}, then the \cond{a} conditions with \cond{b} and \cond{c}, i.e.\ a staircase consistent with the hierarchy.}\label{fig:upset}
\end{figure*}

\begin{figure*}[h]\centering
\includegraphics[width=\linewidth]{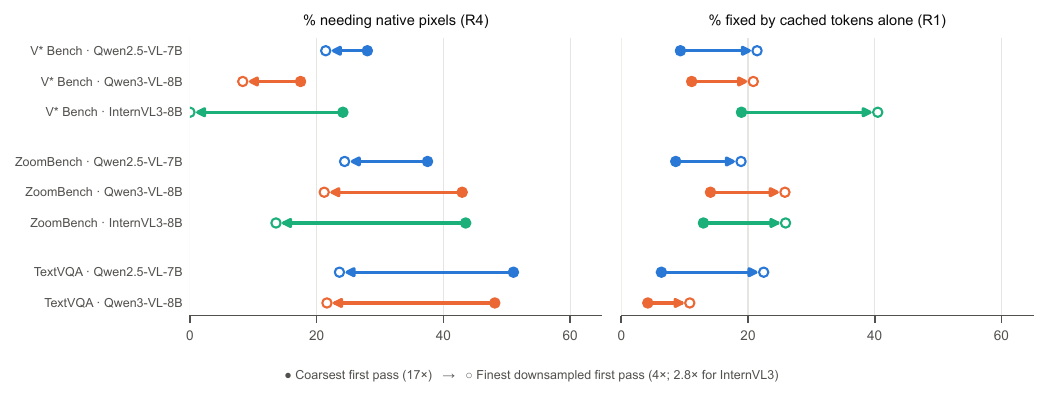}
\caption{\textbf{Per-model change between the coarsest and the finest downsampled first pass.} Filled: 17$\times$, hollow: 4$\times$ (2.8$\times$ for InternVL3). Left: share of zoom-rescued failures that need new pixels (\rfour). Right: share fixed by cached tokens alone (\rone).}\label{fig:dumbbell}
\end{figure*}

\begin{figure*}[h]\centering
\includegraphics[width=\linewidth]{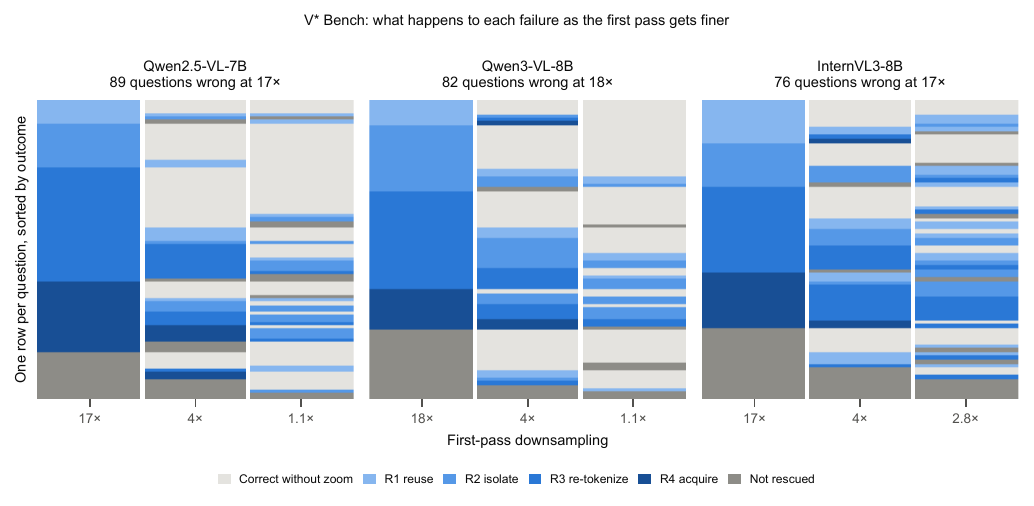}\\
\includegraphics[width=\linewidth]{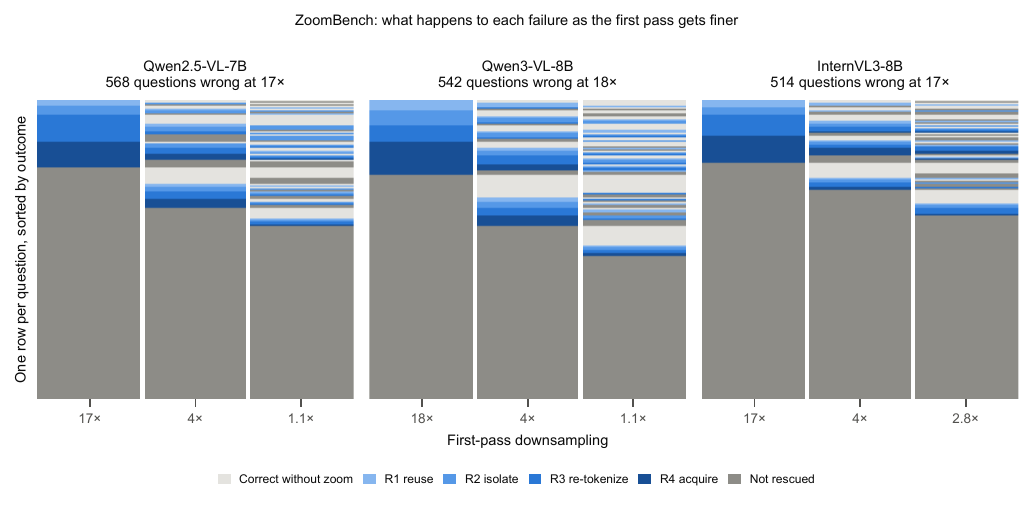}
\caption{\textbf{Per-question trajectories.} Each row is one question whose first-pass answer was wrong at the coarsest budget, columns are first-pass budgets, and color is the outcome (correct without zoom, strict regime \rone--\rfour, or not rescued). Rows are sorted by outcome. As the first pass gets finer most V$^*$ failures become correct outright and \rfour\ empties. On ZoomBench the majority of failures are not fixable by the oracle zoom at any budget.}\label{fig:trajectory}
\end{figure*}

\begin{figure*}[h]\centering
\includegraphics[width=\linewidth]{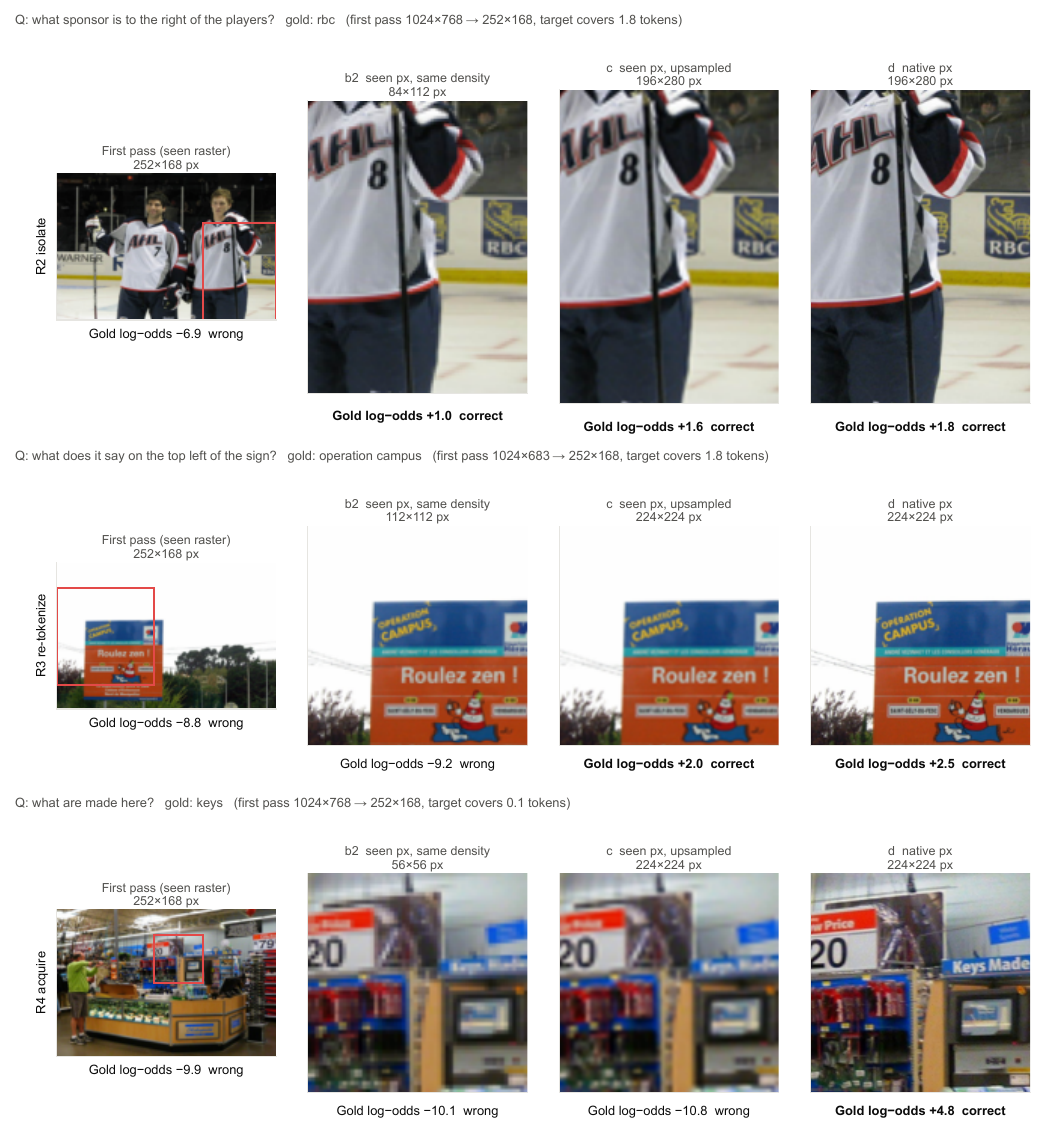}
\caption{\textbf{Three TextVQA failures, one each from \rtwo, \rthree\ and \rfour,} at an 18$\times$ first pass (Qwen2.5-VL-7B): the first pass and the inputs under \cond{b2}, \cond{c} and \cond{d}, with the gold log-odds of each reply.}\label{fig:textvqa_examples}
\end{figure*}

\Cref{fig:examples} shows two more failures in the format of \cref{fig:teaser}, and \cref{fig:textvqa_examples} three TextVQA failures. \Cref{fig:sweep_zb} repeats the sweep of \cref{fig:sweep} on the other two benchmarks. \Cref{fig:forest} gives every condition and its sham with bootstrap intervals, \cref{fig:ladder} the cumulative share fixed as conditions are added in order of cost, \cref{fig:graded} the gold log-odds under each condition, and \cref{fig:upset} the overlap between the sets of failures that the conditions fix. \Cref{fig:dumbbell,fig:trajectory} follow the regimes from the coarsest to the finest budget, per model and per question.
\section{Question subgroups}
\Cref{tab:subgroups} splits the partition by question type (pooled over first passes downsampled $\geq2\times$). Spatial questions on V$^*$ (relative position of two objects) fall into the reuse and isolation regimes more often than attribute questions, whose rescues are dominated by re-tokenization. Counting questions on ZoomBench need new pixels no more often than multiple-choice ones (19--31\% against 30--32\%) but differ in the split between isolation and re-tokenization. All four regimes are populated in every subgroup, and \rthree\ is the largest no-acquisition regime in every subgroup except V$^*$ spatial questions for Qwen3-VL and InternVL3 and counting questions for the two Qwen models, where isolation dominates.

\begin{table*}[h]\centering\footnotesize
\caption{\textbf{Regime partition by question subgroup} (in percent, V$^*$ and ZoomBench, first pass $\geq2\times$ downsampled). The last two columns give the share of each subgroup fixed by upsampling the seen pixels (\cond{c}) and by any cached-token condition.}
\label{tab:subgroups}
\fittable{\begin{tabular}{lllrrrrrrr}
\toprule
Model & Data & Questions & $n$ & \rone & \rtwo & \rthree & \rfour & \cond{c} fixes & \cond{a1}--\cond{a3} fix \\
\midrule
Qwen2.5-VL-7B & V$^*$ & attribute & 90 & 12 & 14 & 48 & 26 & 72 & 14 \\
 & V$^*$ & spatial & 27 & 19 & 26 & 30 & 26 & 81 & 30 \\
 & ZoomBench & multiple choice & 223 & 13 & 19 & 37 & 32 & 63 & 14 \\
 & ZoomBench & counting & 32 & 22 & 34 & 19 & 25 & 62 & 25 \\
\addlinespace[2pt]
Qwen3-VL-8B & V$^*$ & attribute & 87 & 13 & 32 & 43 & 13 & 82 & 16 \\
 & V$^*$ & spatial & 24 & 25 & 46 & 12 & 17 & 75 & 42 \\
 & ZoomBench & multiple choice & 247 & 21 & 23 & 25 & 32 & 58 & 26 \\
 & ZoomBench & counting & 39 & 18 & 31 & 21 & 31 & 62 & 23 \\
\addlinespace[2pt]
InternVL3-8B & V$^*$ & attribute & 111 & 29 & 20 & 38 & 14 & 83 & 30 \\
 & V$^*$ & spatial & 33 & 27 & 39 & 27 & 6 & 91 & 61 \\
 & ZoomBench & multiple choice & 237 & 22 & 17 & 32 & 30 & 66 & 29 \\
 & ZoomBench & counting & 43 & 16 & 19 & 47 & 19 & 77 & 19 \\
\addlinespace[2pt]
\bottomrule
\end{tabular}
}
\end{table*}

\end{document}